\documentclass{article}
\pdfoutput=1
\usepackage{spconf,amsmath,graphicx}
\usepackage{amssymb,amsfonts}
\usepackage{dsfont}
\usepackage{bbm}

\usepackage{gensymb}
\usepackage{algorithm}
\usepackage{algpseudocode}
\algrenewcommand\algorithmicrequire{\textbf{Input:}}
\algrenewcommand\algorithmicensure{\textbf{Output:}}
\usepackage{graphicx}
\usepackage{textcomp}
\usepackage{booktabs}
\usepackage{subcaption}
\usepackage{xcolor}
\usepackage{cite}
\usepackage{booktabs}
\usepackage{subcaption}
\usepackage{textcomp}
\usepackage{url}
\usepackage{relsize}
\usepackage{multirow}

\definecolor{onlineBlue}{RGB}{172, 197, 255}
\definecolor{onlineRed}{RGB}{255, 177, 177}
\usepackage[hidelinks]{hyperref}
\hypersetup{
    colorlinks=false,
    linkcolor=black,
    filecolor=black,
    citecolor=black,
    urlcolor=black,
}

\definecolor{onlineBlue}{RGB}{172, 197, 255}
\definecolor{onlineRed}{RGB}{255, 177, 177}
\definecolor{tangelo}{rgb}{0.98, 0.3, 0.0}
\definecolor{applegreen}{rgb}{0.55, 0.71, 0.0}
\definecolor{unmellowyellow}{rgb}{1.0, 1.0, 0.4}
\usepackage{enumitem}
\setlist{nosep, leftmargin=14pt}

\usepackage{mwe} 
\usepackage{atbegshi}
\AtBeginDocument{\AtBeginShipoutNext{\AtBeginShipoutDiscard}}

\title{Self-Attentive Spatial Adaptive Normalization for Cross-Modality Domain Adaptation}
\name{Devavrat Tomar\textsuperscript{1}, Manana Lortkipanidze\textsuperscript{1}, Guillaume Vray\textsuperscript{1}, Behzad Bozorgtabar\textsuperscript{1,2}, Jean-Philippe Thiran\textsuperscript{1,2}}
\address{\textsuperscript{1} Signal Processing Laboratory (LTS5), \'Ecole Polytechnique F\'ed\'erale de Lausanne (EPFL), Switzerland\\
\textsuperscript{2} CIBM Center for Biomedical Imaging, Switzerland}

\begin{document}
%
\thanks{Devavrat Tomar \& Manana Lortkipanidze contributed equally. }

\maketitle
\begin{abstract}
Despite the successes of deep neural networks on many challenging vision tasks, they often fail to generalize to new test domains that are not distributed identically to the training data. The domain adaptation becomes more challenging for cross-modality medical data with a notable domain shift. Given that specific annotated imaging modalities may not be accessible nor complete. Our proposed solution is based on the cross-modality synthesis of medical images to reduce the costly annotation burden by radiologists and bridge the domain gap in radiological images. We present a novel approach for image-to-image translation in medical images, capable of supervised or unsupervised (unpaired image data) setups. Built upon adversarial training, we propose a learnable self-attentive spatial normalization of the deep convolutional generator network's intermediate activations. Unlike previous attention-based image-to-image translation approaches, which are either domain-specific or require distortion of the source domain’s structures, we unearth the importance of the auxiliary semantic information to handle the geometric changes and preserve anatomical structures during image translation. We achieve superior results for cross-modality segmentation between unpaired MRI and CT data for multi-modality whole heart and multi-modal brain tumor MRI (T1/T2) datasets compared to the state-of-the-art methods. We also observe encouraging results in cross-modality conversion for paired MRI and CT images on a brain dataset. Furthermore, a detailed analysis of the cross-modality image translation, thorough ablation studies confirm our proposed method’s efficacy.

\end{abstract}

\section{Introduction}\label{sec:introduction}
Diagnostic radiology encompasses diverse imaging modalities for disease diagnosis and treatment. The most popular medical imaging modalities used for radiotherapy treatment planning and segmentation of tumor volumes are magnetic resonance imaging (MRI) and computed tomography (CT). MRI has been frequently used by radiologists to aid organ-at-risk delineation due to its superior soft-tissue contrast \cite{westbrook2018mri}; however, compared with CT, it is relatively time-consuming, expensive, and not readily accessible. On the other hand, a CT scan uses X-ray beams to create fine slices \cite{chartrand2007coronary}, which is a useful means of obtaining various levels of detail in dense areas of the body. Nevertheless, CT scans cannot easily distinguish soft tissue with poor contrast. As such, radiologists may opt for an MRI. In practice, due to some limitations such as inadequate scanning time, resources, and cost, certain imaging modalities may be accessible nor complete.

\begin{figure}[!t]
\centering
\begin{tabular}{@{}c@{}c@{}c@{}c@{}c@{}}
     \includegraphics[width=0.2\linewidth]{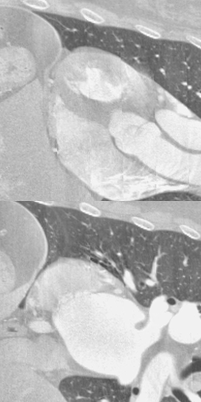} & \includegraphics[width=0.2\linewidth]{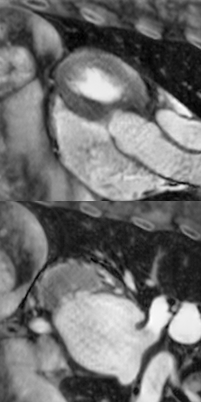} & \includegraphics[width=0.2\linewidth]{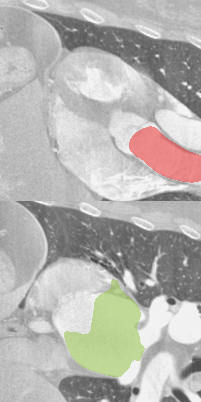} & \includegraphics[width=0.2\linewidth]{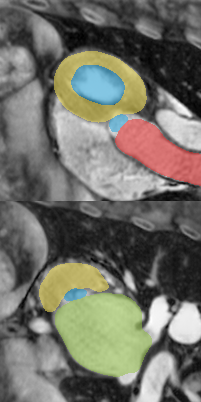} & \includegraphics[width=0.2\linewidth]{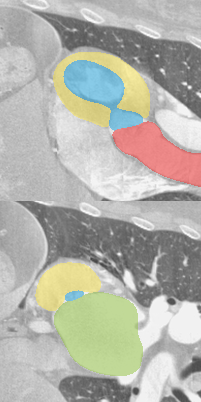}\\
     (a) & (b) & (c) & (d) & (e) 
\end{tabular}

\caption{Examples of unsupervised cross-domain adaptation between MRI/CT for cardiac substructure segmentation: (a) original CT, (b) synthesized MRI, (c) segmentation result on the CT image using a detached U-Net model trained on MRI images, (d) segmentation result on the fake MRI (synthesized from CT) using our proposed SASAN, and (e) ground-truth segmentation.}
\label{fig:introduction_figure}
\vspace{-5mm}
  \end{figure}

These limitations could be overcome by Computer-Aided Diagnosis (CAD) systems that can be deployed to impute the missing image modalities by using other accessible modalities. However, it has been pointed out that established machine learning models would under-perform when tested on data from different modalities due to the domain shift. This problem is more pronounced as the domain shift between MRI/CT data is remarkable such as the difference in the appearance of tissues and anatomical structures. Therefore, there is a strong clinical need to develop cross-modality image translation systems to generalize the learned models from one modality into another. These methods should ideally transfer knowledge across image domains (modalities) without using additional annotations from the target domain. Cross-modality image translation is defined as learning a function $f:\mathcal{X}\to \mathcal{Y}$ that maps images from source domain $\mathcal{X}$ to target domain $\mathcal{Y}$ by using supervision (e.g. paired data) \cite{isola2017image, zhang2018self} or by using unpaired data in an unsupervised setup \cite{CycleGAN2017,kim2017learning}. Image translation methods \cite{bozorgtabar2019learn,CycleGAN2017,bozorgtabar2020exprada,bozorgtabar2019using,mahapatra2019progressive} have gained attention due to the impressive success of conditional generative models such as conditional generative adversarial networks (cGANs) \cite{isola2017image, zhang2018self}. Currently, a large portion of medical image synthesis approaches \cite{nie2017medical, frid2018gan} requires a set of paired samples to build cross-modality reconstruction. However, collecting a large number of aligned image pairs is typically difficult, and sometimes even impractical to obtain. Additionally, if the image pair examples from the two domains are not available, finding such a mapping function is ill-posed as it involves estimating the joint distribution of the two domains from the marginals. Nonetheless, if we assume some prior over the joint-distribution and constraint the family of mapping functions (e.g., similar colors, shapes, cyclic consistency, consistent semantic layouts), we can estimate the mapping function that best matches the given priors and constraints. For example, the idea of the cyclic consistent generative adversarial network (CycleGAN) \cite{CycleGAN2017} performs image translation using only the marginal distributions. However, for a superior quality medical image modality translation, e.g., CT-based MR image construction, we require the organs' anatomical structures to be preserved for detecting unhealthy tissues and other anomalies in the synthesized translated image.

In this paper, we propose SASAN, short for the \textbf{S}elf-\textbf{A}ttentive \textbf{S}patial \textbf{A}daptive \textbf{N}ormalization method, a new medical image modality conversion. We introduce a self-attention module that attends to different anatomical structures of the organ in the image to improve the image translation task. Our method can leverage the auxiliary semantic segmentation information to guide the attention network and be used for paired and unpaired datasets. Our cross-modality domain adaptation has been further validated for medical image segmentation in the absence of target domain labels. Examples of segmentation results for cardiac substructures are shown in Fig. \ref{fig:introduction_figure}. 

\subsection{Contributions}
Our contributions are as follows:

 \begin{itemize}
\item We propose a new plug-and-play framework for unsupervised cross-modality adaptation of the medical image segmentation. Our approach detaches the segmentation network from the cross-modality adaptation process and eliminates the target domain's labeling cost. This is more practical than the earlier approaches, which assume there is access to a few amounts of annotated data on the target domain or train the joint segmentation-domain adaptation models,

\item  We propose a learnable self-attentive spatial normalization of the deep convolutional generator network's intermediate activations to preserve anatomical structures during image translation,

 \item Our proposed attention regularization loss ensures learning orthogonal attention maps, each of them focusing on specific anatomical regions and avoiding redundancy, facilitating the translation process for different anatomical structures. Besides, the proposed auxiliary semantic loss term empowers our method to generate a missing modality with anatomical details, further guaranteeing the effectiveness of our approach towards clinical applications and its generalizability,

 \item We demonstrate the effectiveness of our proposed SASAN in tasks, including unsupervised image translation between multi-modal brain tumor MRI (T1/T2) and cardiac substructure segmentation. The proposed method achieves state-of-the-art performance on the multi-modality whole heart segmentation task in qualitative and quantitative evaluations, making it a potential solution to make accurate clinical decisions.

  \end{itemize}

\section{Related Work} \label{related_work}
Traditional medical image synthesis methods are often divided into three categories: learning-based methods \cite{huynh2015estimating,jog2013magnetic,torrado2016fast}, tissue-segmentation-based methods \cite{zheng2015magnetic,su2015generation}, and atlas-based methods \cite{catana2010toward,arabi2016atlas,ahunbay2019technique}. However, there are certain pros and cons of using these approaches. The learning-based approaches directly build a relationship between image domains by mapping between feature representation of two different domains. Therefore, the quality of mapping is greatly affected by the feature selection process, which is not desirable. Tissue-segmentation based approaches, for instance, require manual refinement of segmentation classes after the segmentation of image voxels into a small number of tissue types. Furthermore, atlas-based approaches use atlas images to align the image and estimate matching pairs in another domain. However, finding an atlas image is not straightforward and poses difficulties in the workflow. Besides, it is hard to cover anatomical geometry variations using atlas data.

The success of Convolutional Neural Networks (CNNs) for the computer vision domain soon got adopted by researchers for biomedical imaging tasks. It alleviates the need for manual representation learning, as in learning-based classical approaches, and offers automation of embeddings. However, training CNNs requires a large amount of annotated data, representing a considerable challenge in the biomedical domain. Besides, biomedical images are scarce due to patients' privacy and require expert knowledge for proper annotation or alignment. Recently, CNNs have been exploited for medical image synthesis \cite{sivaswamy2018retinal}, achieving superior performance to the classical machine-learning methods. For instance, Han \textit{et al}. \cite{han2017mr} utilized CNN for image-modality conversion to synthesize CT images from MR images using paired data. Similarly, Zhao \textit{et al}. \cite{zhao2017whole} proposed a modified U-Net to synthesize an MR image from a CT counterpart by minimizing voxel-wise differences between images. However, these approaches suffer from the problem of blurred outputs caused by the optimization of voxel-wise loss. Therefore, GANs have emerged as state-of-the-art approaches for the domain for their realistic outputs, including the recent variations conditional GAN, e.g., Pix2Pix \cite{isola2017image} and CycleGAN \cite{CycleGAN2017}. For example, Nie \textit{et al}. \cite{nie2017medical} proposed a context-aware GAN to synthesize CT images from input MR images. Bi \textit{et al}. \cite{bi2017synthesis} proposed a GAN-based method built upon the pix2pix framework to synthesize positron emission tomography (PET) images from CT images. There are other concurrent GAN-based methods \cite{zhou2020hi,yu2019ea,zhang2019skrgan,dar2019image}, which have been proposed to learn a transformation in the pixel space for medical synthesis. However, most of these methods use the rigidly aligned cross-modality data and collecting the aligned paired training data is usually expensive. Especially the cross-modality data in medical institutions are often unpaired due to restricted medical conditions. Some recent methods \cite{wolterink2017deep,kim2017learning} alleviate this constraint and impose a cycle consistency condition to use the unpaired data. For instance, Wolterink \textit{et al}. \cite{wolterink2017deep} applied a CycleGAN when synthesizing an MR image using a CT image. However, weak-supervision without additional constraints could lead to higher distortion and inaccurate estimation of details, e.g., soft tissues in output images. For example, the translated images could be heavily skewed towards source image modalities in the CycleGAN. Some other approaches \cite{johnson2016perceptual,gatys2016image} use perceptual loss based on similarity measure in feature space extracted from pre-trained classification networks.  Although these methods show promising results and could generate high-quality synthetic images, they do not incorporate semantic information, limiting their capability to estimate the contextual information during the cross-modality conversion.

All aforementioned GAN-based image translation methods focus more on the overall appearance of the translated image and have difficulty in generating fine structures. Recently, some attention based approaches focus on translating foreground only \cite{mejjati2018unsupervised} or distort the spatial context of the structures \cite{kim2019u}. For example, Kim \textit{et al}. \cite{kim2019u} proposed a method built upon an attention mechanism's idea to focus on the most discriminative image regions that distinguish between source and target domains during image translation. However, their learned attention maps can deform the original structures present in the domain source. This is usually beneficial for specific datasets that require distortion of the source domain's structures for pleasing results. Nonetheless, for medical cross-modality translation, preserving the anatomical structures is very crucial. 

Some recent methods \cite{chen2020unsupervised,saito2018maximum,dou2018pnp} have been proposed to extend image synthesis to a general-purpose unsupervised domain adaptation (UDA) framework. These methods aim to bridge the gap between two domains at different representation levels, such as feature \cite{ganin2016domain,bozorgtabar2019syndemo,luo2017label,saito2018maximum} or pixel levels \cite{bousmalis2017unsupervised} or both image and feature levels \cite{dou2018pnp}, \cite{Hoffman2018cycada} to validate further the quality of the synthesis results for downstream tasks such as segmentation. Qi \textit{et al}. \cite{dou2018pnp} proposed a plug-and-play domain adaptation method that simultaneously matches feature and output distribution across domains. However, the alignment of feature-level marginal distributions does not explicitly enforce semantic coherence between modalities. Instead, we address this issue via learning orthogonal attention maps, each focusing on specific anatomical regions, facilitating the translation process across image modalities. Hoffman \textit{et al}. \cite{Hoffman2018cycada} proposed CyCADA, an adversarial training approach that benefits from both image-level and output-level adaptation for the task of image segmentation. Cai \textit{et al}. \cite{cai2019towards} utilized a cycle consistent generative adversarial network for cross-modal organ translation and segmentation. Zhang \textit{et al}. \cite{zhang2020collaborative} proposed a collaborative UDA method focusing on hard-to-transfer samples to tackle label noise.

Another important track of domain adaptation for image segmentation is based on self-training schema. Zou \textit{et al}. \cite{zou2018unsupervised} proposed a UDA framework based on an iterative self-training procedure for semantic segmentation. They also introduced a class-balanced self-training mechanism to address the class imbalance issue for transfer learning. Xia \textit{et al}. \cite{xia2020uncertainty} proposed an uncertainty-aware multi-view co-training (UMCT) approach to utilize unlabeled data aimed at domain adaptation and semi-supervised learning simultaneously for 3D volumetric image data.

\section{Methods} \label{methods}
 Upon the motivation mentioned in Section \ref{related_work} for the attention-based methods, this paper proposes an unsupervised cross-modality domain adaptation for medical images to focus on preserving the geometrical relationship between the anatomical structures during image translation by utilizing the semantic information of image modalities. 

\begin{figure*}[ht]
  \centering
    \begin{tabular}{@{}c|c@{}}
      \includegraphics[width=0.5\linewidth]{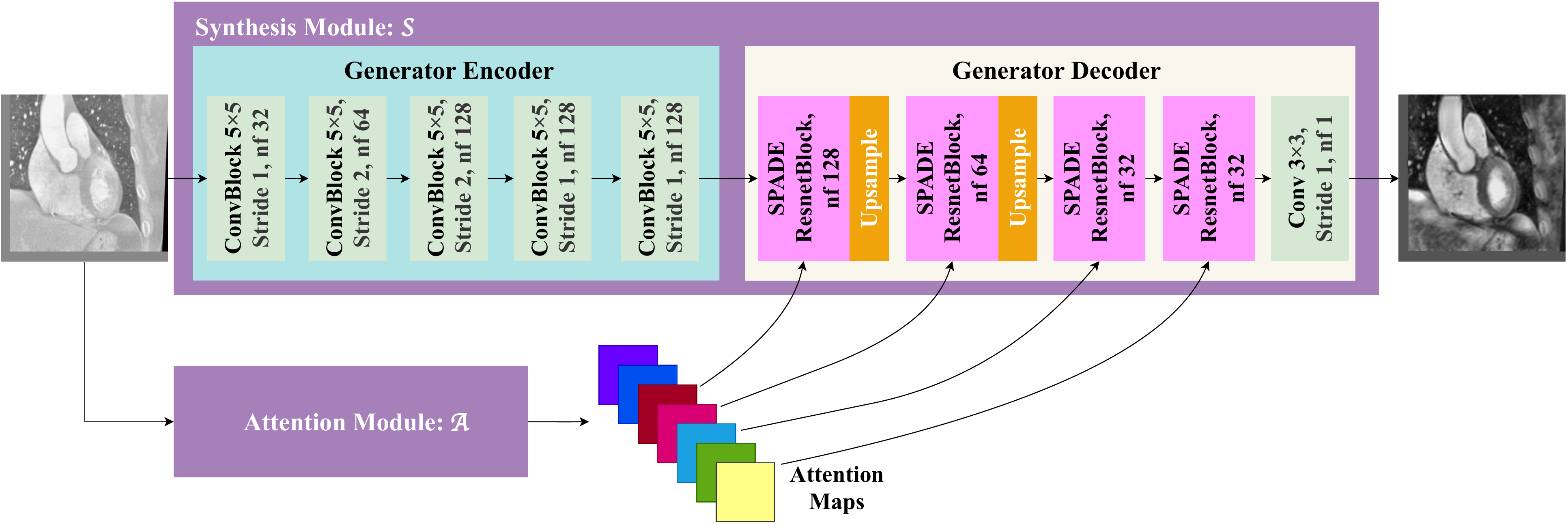} &
      \includegraphics[width=0.22\linewidth]{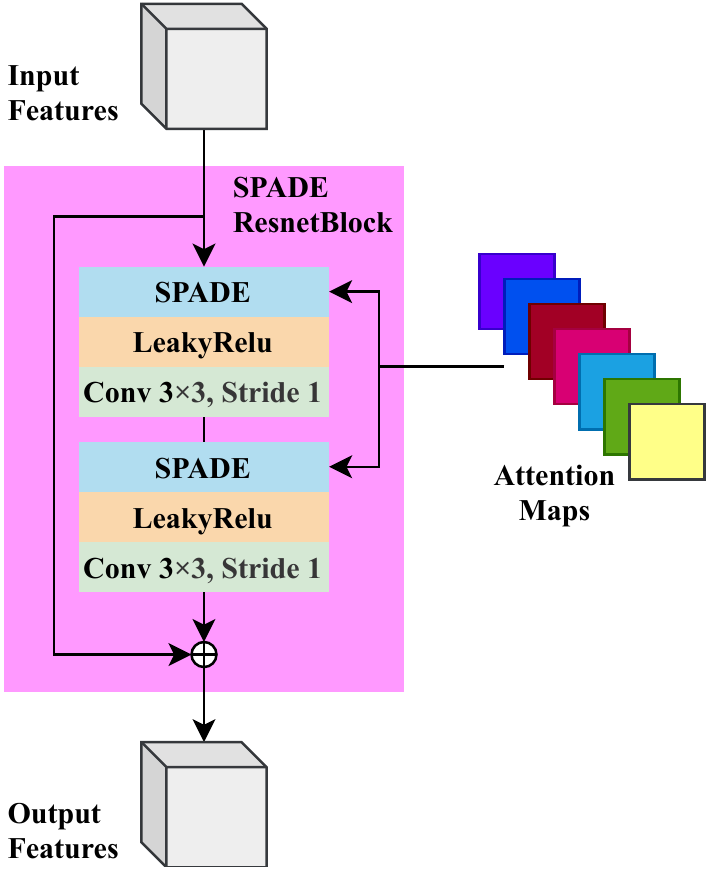}
      \includegraphics[width=0.25\linewidth]{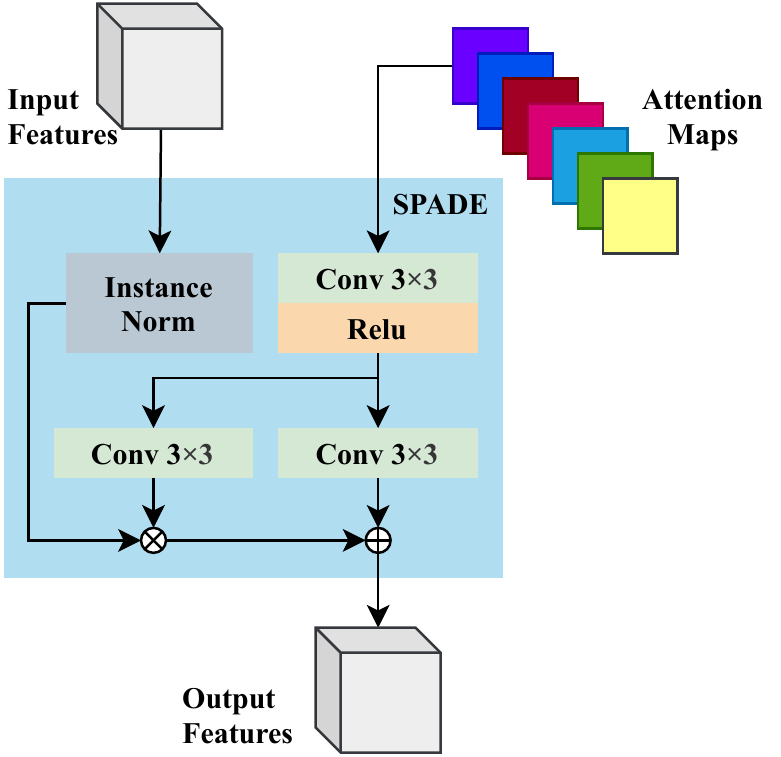}\\
      (a) & (b)
      \end{tabular}
  \caption{\raggedright Illustration of the proposed pipeline. (a) Generator architecture with a self-attention module, and (b) the SPADE ResnetBlock architecture with the proposed self-attentive adaptive normalization.}
\label{fig:generator}
\vspace{-4mm}
\end{figure*}

For an unsupervised cross-modality adaptation, translating different semantic regions between the two corresponding domains is often very important, especially for those biomedical datasets. Different tissue regions should be translated differently from one domain to another for modality conversion between MRI and CT. To achieve this, we propose self-attentive semantically learnable spatial feature normalization of the activations of the generator model, as shown in Fig. \ref{fig:generator}. This normalization technique is similar to \cite{park2019semantic}; however, instead of using the segmentation maps for conditional image generation, we use the proposed learned attention maps to normalize the generator's activations. Our generator can be divided into two modules: synthesis module $\mathbf{\mathcal{S}}$ and a mapping network $\mathbf{\mathcal{A}}$ called attention module. $\mathbf{\mathcal{A}}$ learns the spatial semantic information as attention maps, which are then fed to the decoder layers of $\mathbf{\mathcal{S}}$ to re-normalize their output activations based on the attended information. We also utilize two variants of PatchGAN \cite{CycleGAN2017} based discriminators; one focuses on the image translation quality while the other focuses on the predicted segmentation quality for the two domains. Using a separate discriminator for predicting segmentation labels is motivated by the fact that anatomical structures' spatial relationships do not change for the two domains' images under image translation.

If the segmentation labels are provided e.g., for cardiac structures segmentations \cite{zhuang2013challenges}, we use the segmentation labels to guide the attention mechanism for image-to-image translation based on unpaired data and demonstrate how extra information available can be utilized during training. For the domain adaptation task, we omit the ground truth segmentation labels of one domain (CT) during training and evaluate the performance of a U-Net based MRI segmentation model on the synthesized MRI. For the supervised image translation task, e.g., on brain dataset \cite{rirewebsite}, we also include few pairwise pixel losses, as discussed in Section \ref{sec:loss}, in addition to cycle consistency assumption. As mentioned in \cite{jin2019deep}, including cycle consistency for paired cross-domain data helps to preserve the structures while translating from one domain to another.

\begin{figure*}
    \centering
    \includegraphics[width=\linewidth]{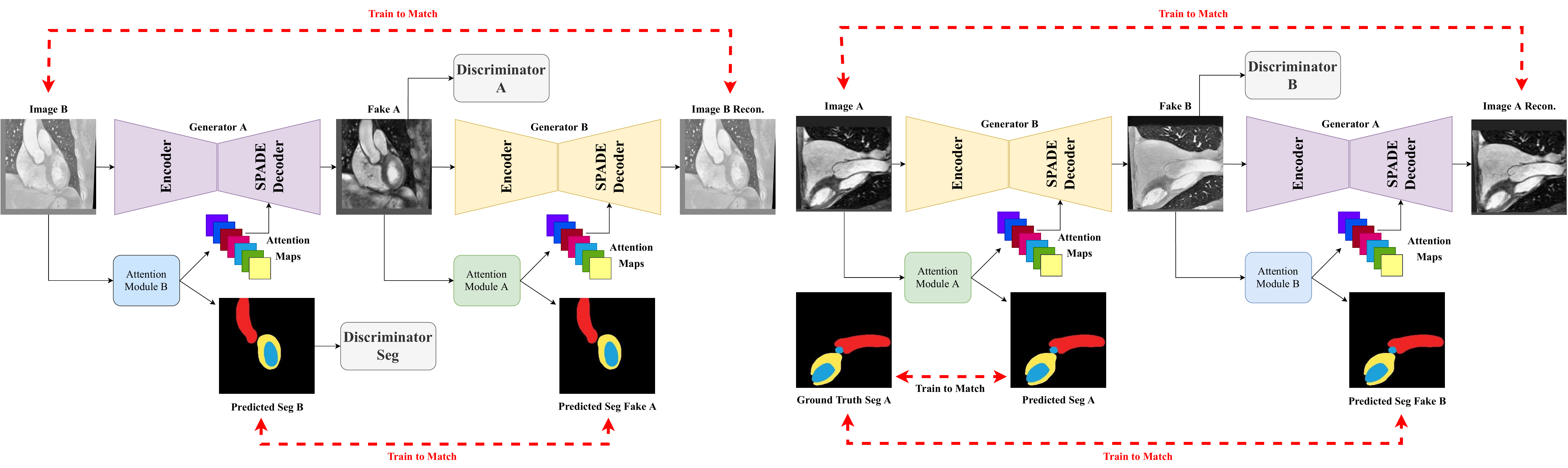}
    \caption{\textbf{Overview of training.} Domain adaptation between MRI (A) and CT (B). Attention Module extracts orthogonal attention features that are fed to the decoder of the generator (SPADE Decoder) for image translation that works on three different resolutions of the attention maps ($64\times64$, $128\times128$, and $256\times256$). Since only the ground truth of domain A is available, we train Attention Module B using an additional adversarial loss on the segmentation labels.}
    \label{fig:training_procedure}
\end{figure*}

\subsection{Self-Attentive Generator Architecture}
As mentioned previously, our generator is composed of two fully convolutional neural networks - synthesis module ($\mathbf{\mathcal{S}}$) and attention module ($\mathbf{\mathcal{A}}$). $\mathbf{\mathcal{A}}$ helps in transmitting the semantic layout information to the intermediate layers of $\mathbf{\mathcal{S}}$, thus making the translation task easier to learn. The architecture of $\mathbf{\mathcal{S}}$ is similar to the Pix2Pix generator architecture \cite{CycleGAN2017}, except that the Encoder is composed of Convolutional Blocks while the Decoder contains SPADEResblock module proposed in \cite{park2019semantic} that operates on different resolutions of attention maps derived from the Attention Module (see Fig. \ref{fig:training_procedure}).


\subsubsection{Attention Module}

    The architecture of the attention module, $\mathcal{A}$, is based on a lighter version of U-Net. We introduce auxiliary losses and a regularization loss term as described in Section \ref{sec:loss} in the training objective to make the attention maps semantically informative and orthogonal to each other.

\subsubsection{Self-Attentive Spatial Adaptive Normalization}
For normalizing the features of the decoder of the synthesis module, the attention maps are first projected onto a smaller embedding space and then convolved to produce affine parameters $ \gamma $ and $ \beta $, as shown in Fig. \ref{fig:generator} (b). Unlike other conditional normalization methods e.g., AdaIN \cite{huang2017arbitrary} and AdaLIN \cite{kim2019u}, $\gamma$ and $\beta$ are not vectors, but tensors with spatial dimensions. The generated $ \gamma $, and $ \beta $ are multiplied and added to the instance normalized \cite{ulyanov2016instance} input features give the final output features:

\begin{align*}
    \hat{x}_{tcij}(m) &= \gamma_{tcij}(m)*\frac{x_{tcij} - \mu_{tc}}{\sigma_{tc}} + \beta_{tcij}(m)\\
    \mu_{tc} &= \frac{1}{HW}\sum_{i,j}x_{tcij}, \;  \sigma^2_{tc} = \frac{1}{HW}\sum_{i,j}(x_{tcij} - \mu_{tc})^2
\end{align*}

where $x$ represents the input features and $\hat{x}(m)$ is the normalized feature based on the attention maps $m$. $\mu_{tc}$ and $\sigma^2_{tc}$ are instance normalized mean and variance of the channel $c$ and batch $t$ over the spatial location $(i, j)$. $H$ and $W$ are height and width of spatial dimensions.

\subsection{Loss Functions} \label{sec:loss}
Our model's full objective comprises several loss functions, some of which are included in the supervised setting when ground-truth is available. We build our model based on the GAN with a min-max objective. Given the images from source domain $\mathcal{X}$, we train a generator $G_{\mathcal{Y}}$ to transform the images from source domain $\mathcal{X}$ to target-like images whose visual appearances are similar to the real images from target domain $\mathcal{Y}$. At the same time, the attention module of $G_{\mathcal{Y}}$ helps in preserving the original structural information. For the unsupervised setting, the discriminators of the domain $\mathcal{Y}$, including the image discriminator $D_{\mathcal{Y}}$ and segmentation discriminator $D_{seg}$ compete with the generator to correctly differentiate the fake outputs from real ones. The attention module for domain $\mathcal{X}$ is denoted as $\mathcal{A}_{\mathcal{X}}$, which acts on the images from domain $\mathcal{X}$. Thus, the output of generator $G_{\mathcal{Y}}(x)$ can be written as $\mathcal{S}_{\mathcal{Y}}(x, \mathcal{A}_{\mathcal{X}}(x))$, where $\mathcal{S}_{\mathcal{Y}}$ is the synthesis module. We also include a single convolution layer $A^{\mathcal{X}}$ on top of the attention module $\mathcal{A}_{\mathcal{X}}$ to get the segmentation predictions on the real and fake images of the domain $\mathcal{X}$. Similarly, we denote the discriminators of domain $\mathcal{X}$ as $D_{\mathcal{X}}$ for images, the generator from domain $\mathcal{Y}$ to $\mathcal{X}$ as $G_{\mathcal{X}}$ and attention module for domain $\mathcal{Y}$ as $\mathcal{A}_{\mathcal{Y}}$. In the supervised settings, we omit the discriminator for segmentation if no segmentation labels are available.

\subsubsection{Adversarial Loss}
We use an adversarial loss to match the distribution of the translated images to the target image distribution. We also employ adversarial loss for matching the distribution of the predicted segmentation labels of translated images and target images to the distribution of target segmentation labels. Using adversarial loss renders more realistic outputs as the generator needs to trick discriminator and make translated images look real. Therefore, the generator tries to minimize the adversarial loss while the discriminator tries to maximize it (detect translated images vs. real images) during training. Instead of using vanilla GAN, we use Least Squares GAN \cite{mao2016squares} objective for stable training:

\begin{equation}
    \mathcal{L}_{lsgan}^{\mathcal{X} \to \mathcal{Y}} = {\mathcal{L}^{Img}_{gan}}^{\mathcal{X} \to \mathcal{Y}} + {\mathcal{L}^{Seg}_{gan}}^{\mathcal{X} \to \mathcal{Y}}
\end{equation}
\vspace{0mm}
where,
\vspace{0mm}
\begin{equation*}
    \begin{split}
        {\mathcal{L}^{Img}_{gan}}^{\mathcal{X} \to \mathcal{Y}} &= \mathrm{E}_{y \sim \mathcal{Y}}\big[D_{\mathcal{Y}}(y)^2\big]
    +\mathrm{E}_{x \sim \mathcal{X}} \big[(1 - D_{\mathcal{Y}}(G_{\mathcal{Y}}(x))^2\big]\\
        {\mathcal{L}^{Seg}_{gan}}^{\mathcal{X} \to \mathcal{Y}} &= 2*\mathrm{E}_{y_{seg} \sim \mathcal{Y}_{seg}}\big[D_{seg}(y_{seg})^2\big]\\
        &+\mathrm{E}_{y \sim \mathcal{Y}} \big[(1 - D_{seg}(A^{\mathcal{Y}}(\mathcal{A}_{\mathcal{Y}}(y)))^2\big]\\
        &+ \mathrm{E}_{x \sim \mathcal{X}} \big[(1 - D_{seg}(A^{\mathcal{Y}}(\mathcal{A}_{\mathcal{Y}}(G_{\mathcal{Y}}(x)))))^2\big]
    \end{split}
\end{equation*}

For domain adaptation, since only segmentation labels for domain $\mathcal{X}$ are available, we replace $y_{seg} \sim \mathcal{Y}_{seg}$ with $x_{seg} \sim \mathcal{X}_{seg}$ in the above equation as the distribution of the two domains' segmentation labels should be the same. The attention module $\mathcal{A}_\mathcal{Y}$ also predicts the segmentation labels of the domain $\mathcal{Y}$, which are discriminated against the segmentation labels of the domain $\mathcal{X}$ using the discriminator. This helps the generator $G_\mathcal{X}$ to generate fake images whose segmentation labels are similar to real segmentation labels by enforcing geometric relationships between the different structures.

\subsubsection{Dual Cycle-Consistency Loss}
Dual cycle-consistency loss helps in avoiding model collapse during training. Given an image $x \in \mathcal{X}$, after the sequential translations of $x$ from domain $\mathcal{X}$ to $\mathcal{Y}$ and $\mathcal{Y}$ to $\mathcal{X}$, the image must be reconstructed back. The dual cycle-consistent loss and adversarial loss play complementary roles. The former encourages a tight relationship between domains, while the latter helps generate realistic images in an unsupervised training. To avoid over-fitting, we furthermore use Structural Similarity Loss (SSIM) and $L_1$ loss for penalizing the cyclic reconstruction:

\begin{equation}
    \begin{split}
        \mathcal{L}_{cycle}^{\mathcal{X} \to \mathcal{Y} \to \mathcal{X}} &= \mathrm{E}_{x \sim \mathcal{X}}\big[\|G_{\mathcal{X}}(G_{\mathcal{Y}}(x)) - x \|_1\big]\\
        +&\mathrm{E}_{x \sim \mathcal{X}}\big[1 -  \mathcal{L}_{ssim}(G_{\mathcal{X}}(G_{\mathcal{Y}}(x)), x)\big]
    \end{split}
\end{equation}

where,
\begin{equation*}
    \mathcal{L}_{ssim}(a,b) =\sum_{i}\frac{(2\mu_{a}^i\mu_{b}^i + \epsilon_1)(2\sigma_{ab}^i + \epsilon_2)}{({\mu_{a}^{i}}^2 + {\mu_{b}^{i}}^2+\epsilon_1)({\sigma_{a}^{i}}^2+{\sigma_{b}^{i}}^2+\epsilon_2)}
\end{equation*}

for some small $\epsilon_1$ and $\epsilon_2$. ($\mu_a^i$, $\mu_b^i$) and ($\sigma_a^i$, $\sigma_b^i$) represent the means and standard deviations respectively of the $i^{th}$ path in images $a$ and $b$.

\subsubsection{Identity Loss}
We also apply an identity consistency constraint, which can regularize the generator to preserve the colors and intensities during translation:

\begin{equation}
    \begin{split}
        \mathcal{L}_{identity}^{\mathcal{Y} \to \mathcal{Y}} = \mathrm{E}_{y \sim \mathcal{Y}}\big[\|G_{\mathcal{Y}}(y) - y \|_1\big]
    \end{split}
\end{equation}

\subsubsection{Attention Regularization Loss}
We propose an attention regularization loss term that encourages the attention maps to be orthogonal with each other. This ensures that the attention maps learn to attend to different regions in the image by avoiding redundancy and increasing the translation power of the model for different anatomical structures. The loss does not require paired images and can be used in an unsupervised manner omitting the need for extensive segmentation:

\begin{equation}
\begin{split}
    \mathcal{L}^{\mathcal{X}}_{reg} &=\mathrm{E}_{x\sim \mathcal{X}}\big[\| \mathcal{A}_{\mathcal{X}}(x)\mathcal{A}_{\mathcal{X}}(x)^{T} - I\|_{F}\big]\\
    +&\mathrm{E}_{y\sim \mathcal{Y}}\big[\| \mathcal{A}_{\mathcal{X}}(G_{\mathcal{X}}(y))\mathcal{A}_{\mathcal{X}}(G_{\mathcal{X}}(y))^{T} - I\|_{F}\big],
\end{split}
\end{equation}

where $I$ is the identity matrix and $\|.\|_{F}$ denotes the Frobenius norm. 

\subsubsection{Auxiliary Losses}
We observed that training the attention module $\mathcal{A}$ in a purely unsupervised manner is not trivial and very time-consuming. Thus, to speed up the process and improve the attention model's power, we leverage all information provided with the dataset. Therefore, we utilize auxiliary losses that use the segmentation maps of the domain (if available), making the multi-task model depending on the data's availability. Furthermore, we ensure that semantic information present in the attention maps is consistent across domains. To achieve this, we include one layer of convolution over the attention maps to predict the segmentation labels and compare them with the auxiliary ground-truth segmentation if available using cross-entropy (CE) and Dice loss (DSC) on real and fake images. For the case of domain adaptation, we replace the ground-truth segmentation of the source domain (e.g., CT) with pseudo labels derived from the attention module on the fake image (e.g., fake MR):

\begin{equation}
\begin{split}
    \mathcal{L}^{\mathcal{X}}_{aux} &=\mathrm{E}_{x\sim \mathcal{X}}\Big[\sum_c CE(x^{c}_{seg}, A^{\mathcal{X}}_{c}(\mathcal{A}_{\mathcal{X}}(x)))\\
    &+\sum_c DSC(x^{c}_{seg}, A^{\mathcal{X}}_{c}(\mathcal{A}_{\mathcal{X}}(x)))\Big]\\
    +&\mathrm{E}_{y\sim \mathcal{Y}}\Big[\sum_c CE(y^{c}_{seg},A^{\mathcal{X}}_{c}( \mathcal{A}_{\mathcal{X}}(G_{\mathcal{X}}(y))))\\
    &+\sum_c DSC(y^{c}_{seg},A^{\mathcal{X}}_{c}( \mathcal{A}_{\mathcal{X}}(G_{\mathcal{X}}(y)))\Big])
\end{split}
\end{equation}

where,

\begin{equation*}
\begin{split}
CE(a,b)&=-\sum _{i,j}a_{i,j}\log(b_{i,j})\\
DSC(a,b)&=1-\frac{\sum_{i,j}2a_{i,j}b_{i,j}}{\sum_{i,j}a_{i,j}+b_{i,j}}
\end{split}
\end{equation*}
$(i,j)$ represent the pixel spatial location and $A^{\mathcal{X}}_{c}$ is one convolutional layer classifier for segmentation class $c$.

\subsubsection{Voxel-Wise Loss}
If the paired dataset is available, we derive $l_1$ loss to impose a pixel-level penalty between the translated image and the ground-truth. This loss is sensitive to the alignment of images; thus, it can not be applied in an unsupervised manner:

\begin{equation}
    \mathcal{L}_{voxel}^{\mathcal{X} \to \mathcal{Y}} = \mathrm{E}_{x \sim \mathcal{X}, y \sim \mathcal{Y}}\big[\|G_{\mathcal{Y}}(x) - y\|_1\big]
\end{equation}

\begin{figure*}[t]
    \centering
    \begin{tabular}{@{}c@{}c@{}c@{}c@{}c@{}c@{}}
    \includegraphics[width=0.1666\linewidth]{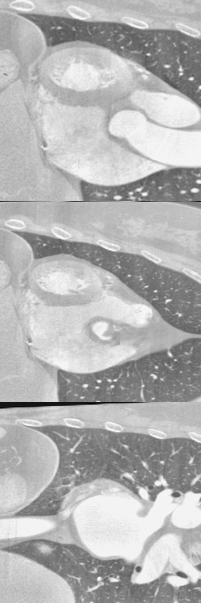} &
    \includegraphics[width=0.1666\linewidth]{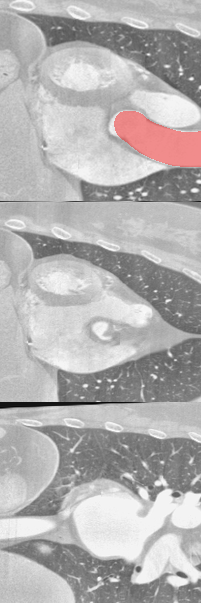} &
    \includegraphics[width=0.1666\linewidth]{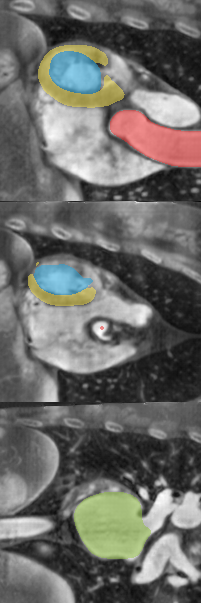} &
    \includegraphics[width=0.1666\linewidth]{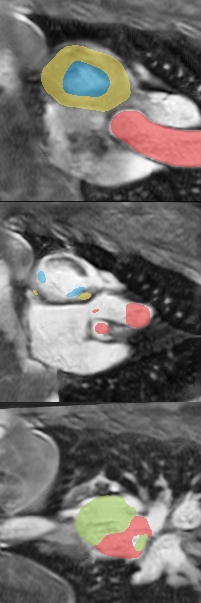} &
    \includegraphics[width=0.1666\linewidth]{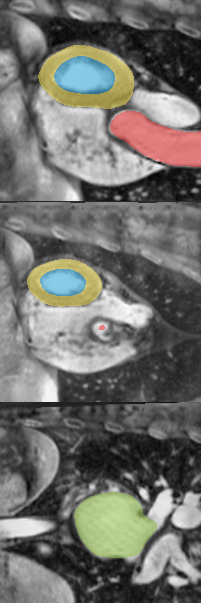} &
    \includegraphics[width=0.1666\linewidth]{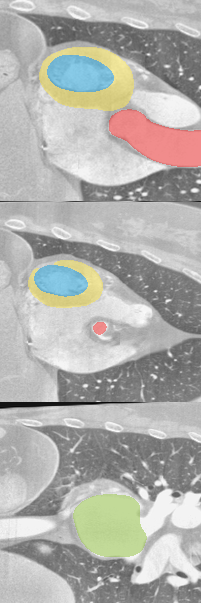} \\
    (a) & (b) & (c) & (d) & (e) & (f)
    \end{tabular}
\caption{Visual comparison of segmentation results produced by different domain adaptation methods between MRI/CT data for the cardiac substructure segmentation task: for each baseline, we evaluate the segmentation accuracy of generated MRI from CT using a detached U-Net segmentation model that was trained on real MRI: (a) original CT, (b) results without domain adaptation, (c) Cycle-GAN (d) U-GAT-IT, (e) SASAN (ours), and (f) ground-truth. The structures of AA, LA-blood, LV-blood and LV-myo are highlighted by \textcolor{onlineRed}{\textbf{red}}, \textcolor{applegreen}{\textbf{green}}, \textcolor{onlineBlue}{\textbf{blue}} and \textcolor{yellow}{\textbf{yellow}} colors, respectively. Best viewed in color.
}
    \label{fig:seg_results}
\end{figure*}

\begin{table*}[h]
\centering
\caption{Performance comparison with different unsupervised cross-domain (MRI $\rightarrow$ CT) adaptation methods for heart structures segmentation. MR-Seg and CT-Seg represent trained U-Net model for MRI and CT domains, which are fine-tuned, and the segmentation results are reported volume-wise.  (* implies as reported in the published paper.) We get an overall 2\% improvement in average Dice score compared to SIFA (p-value$<$0.3 for Welch's t-test on slices.)}
\label{table:domain_adaption_results_heart}
\resizebox{\textwidth}{!}{
\begin{tabular}{@{}|c|cc|cc|cc|cc|cc|@{}}
\toprule
\textbf{Method} & \multicolumn{2}{c|}{\textbf{AA}} & \multicolumn{2}{c|}{\textbf{LA-blood}} & \multicolumn{2}{c|}{\textbf{LV-blood}} & \multicolumn{2}{c|}{\textbf{LV-myo}} & \multicolumn{2}{c|}{\textbf{Mean}} \\
                & \textbf{Dice}  & \textbf{ASSD}   & \textbf{Dice}      & \textbf{ASSD}     & \textbf{Dice}      & \textbf{ASSD}     & \textbf{Dice}    & \textbf{ASSD}     & \textbf{Dice}    & \textbf{ASSD}   \\ \midrule
MR-Seg    & 0.84$\pm$0.05    & 3.6$\pm$2.3      & 0.86$\pm$0.08      & 2.2$\pm$1.1     & 0.92$\pm$0.03      & 2.9$\pm$2.3  & 0.79$\pm$0.03  & 2.8$\pm$1.8        & 0.85$\pm$0.04   & 2.9$\pm$0.5 \\
CT-Seg & 0.86$\pm$0.17  & 5.5$\pm$4.0 & 0.91$\pm$0.02  & 5.2$\pm$0.9 & 0.92$\pm$0.02 & 3.6$\pm$2.3 &  0.86$\pm$0.03  & 5.8$\pm$3.7 & 0.89$\pm$0.03 & 5.0$\pm$0.8 \\
\midrule
W/o adaptation & 0.23$\pm$0.15  & 41.1$\pm$20.5 & 0.13$\pm$0.13  & 30.9$\pm$18.6 & 0.00$\pm$0.00 & N.A. & 0.01$\pm$0.01 & 35.0$\pm$6.7 & 0.09$\pm$0.09 & N.A.\\
CycleGAN  & 0.70$\pm$0.07  & 13.6$\pm$2.8   &  0.69$\pm$0.06 & 11.6$\pm$3.8 & 0.52$\pm$0.20 & 9.3$\pm$3.9   & 0.29$\pm$0.13 & 8.8$\pm$4.2  & 0.55$\pm$0.13  & 10.9$\pm$3.8   \\
U-GAT-IT  & 0.68$\pm$0.08  & 12.0$\pm$3.4 & 0.66$\pm$0.08      & 13.7$\pm$4.1     & 0.55$\pm$0.16      & 8.9$\pm$3.5     & 0.39$\pm$0.12    & 8.9$\pm$3.3     & 0.57$\pm$0.11   & 10.9$\pm$3.5  \\
Pnp-Ada-Net*     & 0.74$\pm$0.07 & 12.8$\pm$3.2  & 0.68$\pm$0.05 & 6.3$\pm$2.3     & 0.62$\pm$0.11      &17.4$\pm$7.0     & 0.51$\pm$0.07    & 14.7$\pm$4.8   & 0.64$\pm$0.08   &12.8$\pm$4.3    \\
SIFA    & 0.84$\pm$0.05 & 7.01$\pm$2.70 & 0.84$\pm$0.04 & 3.8$\pm$1.1 & 0.72$\pm$0.14 & 4.82$\pm$2.33  & 0.62$\pm$0.16 & 4.73$\pm$1.61 & 0.76$\pm$0.08 &  5.1$\pm$1.4  \\
SASAN (final) & 0.82$\pm$0.02 & 4.14$\pm$1.6 & 0.76$\pm$0.25 & 8.3$\pm$2.2 & 0.82$\pm$0.03 & 3.5$\pm$1.1 & 0.72$\pm$0.08 & 3.3$\pm$0.9 & 0.78$\pm$0.10 & 4.9$\pm$1.5 \\ \bottomrule
\end{tabular}}
\end{table*}

\begin{table*}[h]
\centering
\caption{Performance comparison with different unsupervised cross-domain (CT $\rightarrow$ MRI) adaptation methods for heart structures segmentation. Our method beats the second best model SIFA on average Dice score by a margin on 5\% (p-value $<$ 0.2 for Welch's t-test on slices.)}
\label{table:domain_adaption_results_heart_ct_to_mr}
\resizebox{\textwidth}{!}{
\begin{tabular}{@{}|c|cc|cc|cc|cc|cc|@{}}
\toprule
\textbf{Method} & \multicolumn{2}{c|}{\textbf{AA}} & \multicolumn{2}{c|}{\textbf{LA-blood}} & \multicolumn{2}{c|}{\textbf{LV-blood}} & \multicolumn{2}{c|}{\textbf{LV-myo}} & \multicolumn{2}{c|}{\textbf{Mean}} \\
                & \textbf{Dice}  & \textbf{ASSD}   & \textbf{Dice}      & \textbf{ASSD}     & \textbf{Dice}      & \textbf{ASSD}     & \textbf{Dice}    & \textbf{ASSD}     & \textbf{Dice}    & \textbf{ASSD}   \\ \midrule

W/o adaptation  & 0.01 $\pm$0.01 & 45.0$\pm$28.1 & 0.04$\pm$0.07 & 40.2 $\pm$ 10.5 & 0.28 $\pm$0.19 & 17.0 $\pm$ 7.3 & 0.01$\pm$0.01 & 27.9$\pm$6.8 & 0.09 $\pm$0.11 & 32.5$\pm$10.9 \\
CycleGAN    & 0.53 $\pm$0.14 & 17.9$\pm$4.3 & 0.41$\pm$0.08 & 13.3$\pm$4.4 & 0.65$\pm$0.13 & 8.7$\pm$5.3  & 0.44$\pm$0.09 & 6.7$\pm$3.2 & 0.51 $\pm$0.14 & 11.7$\pm$6.0  \\
U-GAT-IT  & 0.55 $\pm$ 0.15  & 16.5$\pm$4.4 & 0.39$\pm$0.10 & 12.1$\pm$3.1 & 0.69$\pm$0.11 & 7.63$\pm$5.2 & 0.49 $\pm$ 0.08 & 7.0 $\pm$3.3 & 0.53$\pm$0.13 & 11.0 $\pm$ 5.3\\
Pnp-Ada-Net* & 0.44$\pm$0.11 & 11.4$\pm$3.2 & 0.47$\pm$0.07& 14.5$\pm$4.1 & 0.78$\pm$0.10 & 4.5$\pm$1.4 & 0.49$\pm$0.03& 5.3$\pm$1.8  & 0.54$\pm$0.08  & 8.9$\pm$2.6 \\
SIFA    & 0.70$\pm$0.04 & 5.0$\pm$2.3 & 0.65$\pm$0.07 & 7.9$\pm$2.8 & 0.78$\pm$0.03 & 4.1$\pm$0.7 & 0.45$\pm$0.03 & 4.4$\pm$0.70 & 0.65$\pm$0.01 &  5.4$\pm$1.0 \\
SASAN (low Res.) & 0.34$\pm$0.09 & 23.3 $\pm$6.4 & 0.58$\pm$0.09 & 12.9$\pm$2.6 & 0.76$\pm$0.07 & 6.8$\pm$2.1 & 0.46$\pm$0.09 & 9.5$\pm$1.6 & 0.53$\pm$0.06 & 13.1$\pm$1.7\\
SASAN (16 Att.) & 0.41$\pm$0.10 & 23.4 $\pm$5.0 & 0.64$\pm$0.10 & 11.1$\pm$2.7 & 0.84$\pm$0.06 & 5.9$\pm$3.4 & 0.64$\pm$0.04 & 3.7$\pm$1.6 & 0.63$\pm$0.05 & 11.1$\pm$5.7 \\
SASAN (final) & 0.54$\pm$0.09 & 18.8 $\pm$5.0 & 0.73$\pm$0.06 & 9.4$\pm$3.0 & 0.86$\pm$0.09 & 6.1$\pm$4.3 & 0.68 $\pm$ 0.08 & 3.85$\pm$1.7 & 0.70$\pm$0.11 & 9.5$\pm$3.2 \\
\bottomrule
\end{tabular}}
\end{table*}





\section{Experiments: Unsupervised Setup}
The effectiveness of our proposed unsupervised cross-domain adaptation method is validated with applications on MRI/CT images for the cardiac structure segmentation (see Fig. \ref{fig:seg_results}) and MRI-T1/T2 for brain tumor segmentation. For the latter task, even though we have access to the paired images, we use them in an unsupervised manner. A U-Net based segmentation model is detached from the learning of our unsupervised domain adaptation to conduct a comprehensive evaluation of different baselines. For each adaptation setting, the ground-truths of target images were used for assessment only, without being used during the training phase. 
\begin{figure*}[t]
    \centering
    \begin{subfigure}[b]{\linewidth}{
    \includegraphics[height=\linewidth, angle=90]{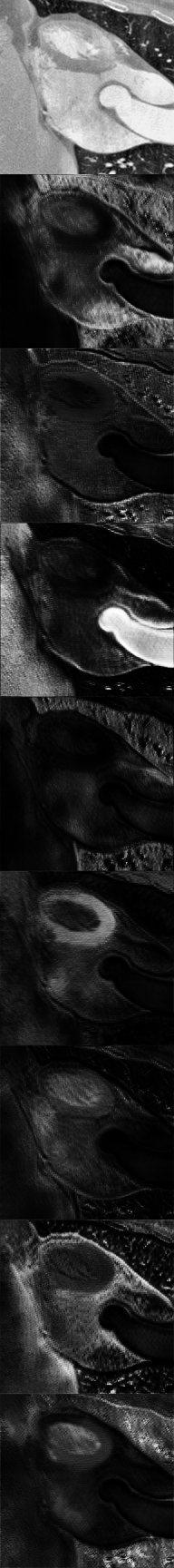}
    \includegraphics[height=\linewidth, angle=90]{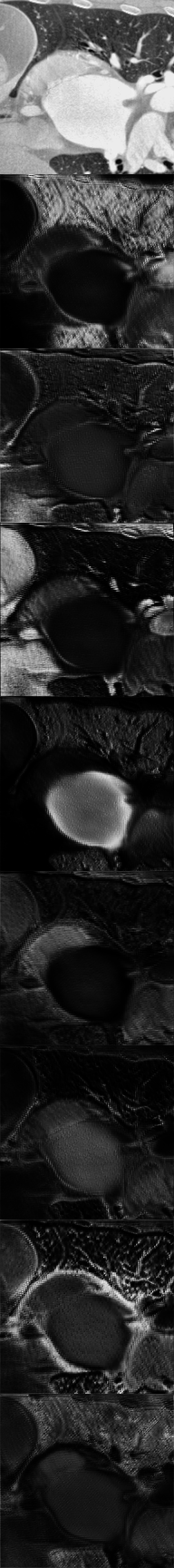}
    \includegraphics[height=\linewidth, angle=90]{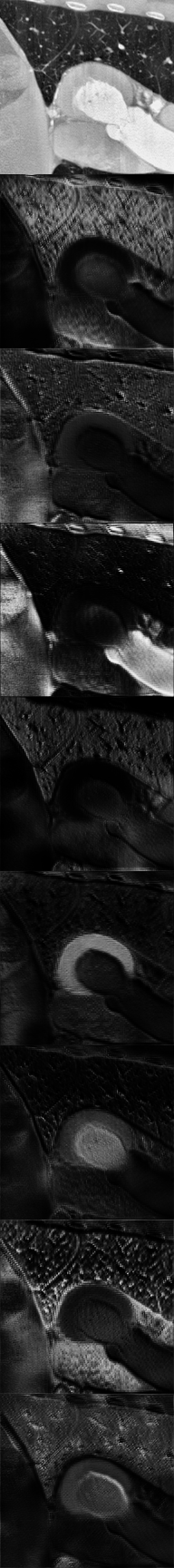}
    \caption{Attention maps on CT images with $\mathcal{L}_{reg}$}
    }
    \end{subfigure}
    
    \begin{subfigure}[b]{\linewidth}{
    \includegraphics[height=\linewidth, angle=90]{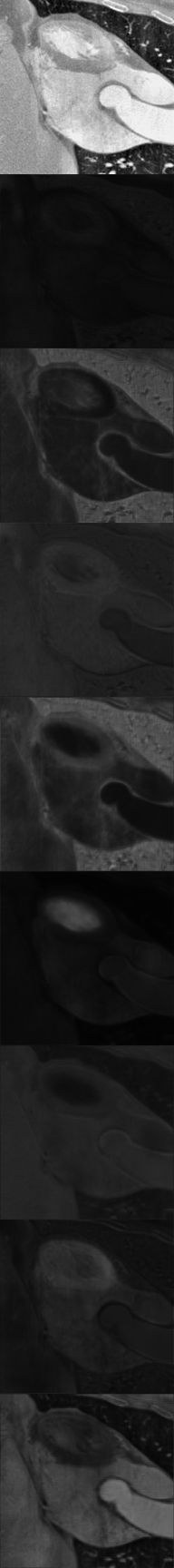}
    \includegraphics[height=\linewidth, angle=90]{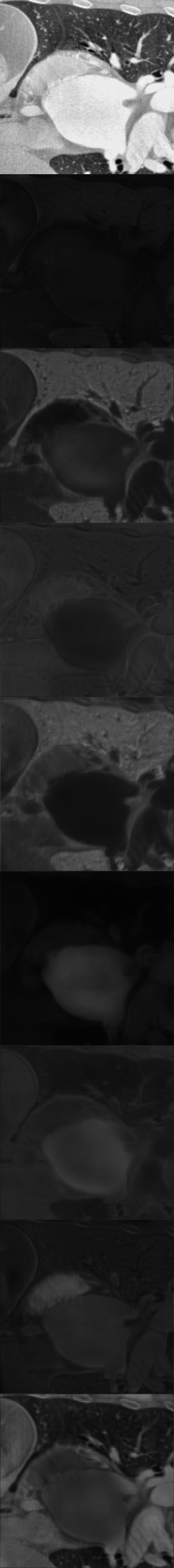}
    \includegraphics[height=\linewidth, angle=90]{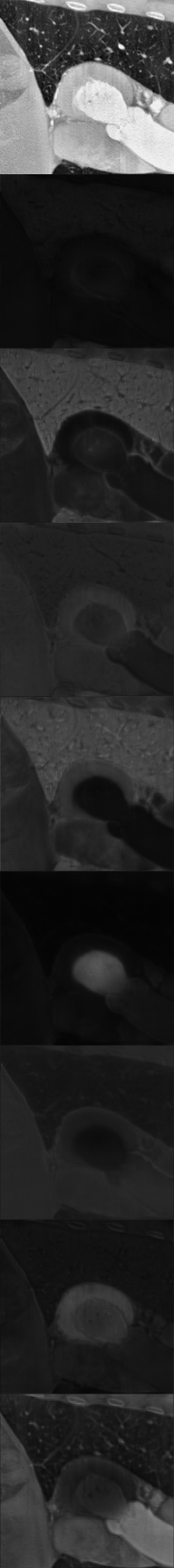}
    \caption{Attention maps on CT images without $\mathcal{L}_{reg}$}
    }
    \end{subfigure}
    \caption{The visualizations of the attention maps for CT cardiac substructures for MRI $\to$ CT domain adaptation. \textit{Left to right:} CT image, corresponding $8$ attention maps.}
    \label{fig:cardiac_attention}
\end{figure*}

\begin{figure}
    \centering
    \begin{tabular}{@{}c@{}c@{}c@{}c@{}}
    \includegraphics[width=0.25\linewidth]{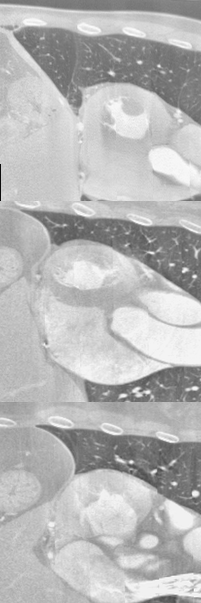} &
    \includegraphics[width=0.25\linewidth]{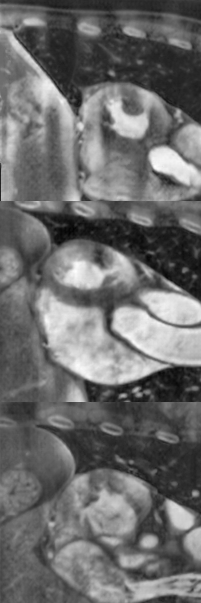} &
    \includegraphics[width=0.25\linewidth]{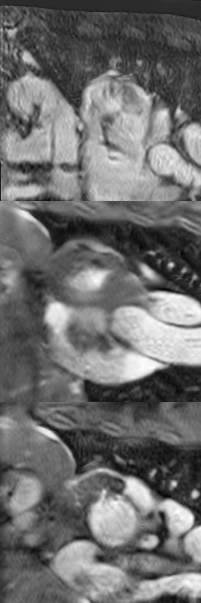} &
    \includegraphics[width=0.25\linewidth]{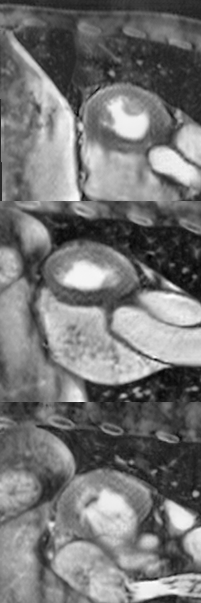}\\
    (a) & (b) & (c) & (d)
    \end{tabular}
    \caption{Qualitative results of different image translation methods from CT to MRI. From left to right: (a) input CT, the translation results of (b) Cycle-GAN, (c) U-GAT-IT, (d) SASAN. }
    \label{fig:image_translation}
    \vspace{-4mm}
\end{figure}

\subsection{Datasets}
For unpaired data, we evaluate our domain adaptation method on MRI-CT whole heart segmentation dataset \cite{zhuang2013challenges} and MRI-T1/T2 brain tumor segmentation dataset \cite{menze2014multimodal} as described below. 

\subsubsection{Whole-heart Segmentation Dataset \cite{zhuang2013challenges}}
This dataset (MICCAI 2017 challenge) contains 20 MRI and 20 CT whole cardiac images with accurate manual segmentation annotations of ascending aorta (AA), left atrium blood cavity (LA-blood), left ventricle blood cavity (LV-blood) and myocardium of the left ventricle (LV-myo). We randomly split each modality of the data into $80\%$ training (16 subjects) and $20\%$ testing (4 subjects) subsets for all experiments, so there is no subject overlap ID among the subsets. For a fair comparison with SIFA approach \cite{chen2020unsupervised}, we used the same preprocessing as the SIFA. We first crop the central heart region for the pre-processing, with four cardiac substructures selected for segmentation. Then, for each 3D cropped image, we perform histogram filtering between 2 and 98 percentile of pixel values followed by normalization to zero-mean and unit standard deviation.

\subsubsection{Multimodal Brain Tumor Segmentation \cite{t1t2dataset}}
The dataset (BRATS, 2015) contains 65 multi-contrast MR scans from glioma patients and consists of four different contrasts - $T1$, $T1c$, $T2$, and FLAIR. Furthermore, expert annotations are given for “edema,” “non-enhancing (solid) core,” “necrotic (or fluid-filled) core,” and “enhancing core” segmentation classes. For the pre-processing, the dataset was co-registered for each subject's image volumes to the T1c MRI, since it had the highest spatial resolution. Additionally, we clipped $2$ and $98$ percentiles to remove outliers. We split the data randomly at the subject-level into on-overlapping training ($90\%$), and test ($10\%$) sets.

\textbf{Data Augmentation.} We use the same data augmentation for all datasets. We perform transformation operations, such as random rotation, random width, and height scaling [0.75, 1], flipping an image horizontally/vertically.

\subsection{Experimental Setup}
For unsupervised settings, we train the two generators ($G_\mathcal{X}$, $G_\mathcal{Y})$ and discriminators ($D_\mathcal{X}$, $D_{seg}$, $D_\mathcal{Y}$) jointly in a cyclic adversarial manner by optimizing the following objective:

\begin{equation}
\min_{G_{\mathcal{Y}}, G_{\mathcal{X}}, \mathcal{A}_c^{\mathcal{Y}}, \mathcal{A}_c^{\mathcal{X}}}\max_{D_{\mathcal{Y}}, D_{\mathcal{X}}, D_{seg}} \mathcal{L}^{\mathcal{X} \to \mathcal{Y}}_{un\_paired} + \mathcal{L}^{\mathcal{Y} \to \mathcal{X}}_{un\_paired}
\end{equation}

where,
\begin{equation*}
    \begin{split}
        \mathcal{L}^{\mathcal{X} \to \mathcal{Y}}_{un\_paired} &= \mathcal{L}_{lsgan}^{\mathcal{X} \to \mathcal{Y}} + \lambda_c \mathcal{L}_{cycle}^{\mathcal{X} \to \mathcal{Y} \to \mathcal{X}}+\lambda_{id}\mathcal{L}_{identity}^{\mathcal{Y} \to \mathcal{Y}}\\
        +&\lambda_{reg}\mathcal{L}^{\mathcal{X}}_{reg} + \lambda_{aux}\mathcal{L}^{\mathcal{X}}_{aux}
    \end{split}
\end{equation*}

We alternatively update the generators and discriminator in one step using Adam optimizer \cite{kingma2014adam} with parameters $\beta_1 = 0.5$, $\beta_2=0.999$ and learning rate of $10^{-4}$ for the first $50$ epochs which is linearly decreased to $0$ for the next $50$ epochs. Fig. \ref{fig:image_translation} shows qualitative comparison of image translation of our method with CycleGAN\cite{CycleGAN2017} and U-GAT-IT \cite{kim2019u}. For evaluating the methods quantitatively, we train a separate U-Net based segmentation model to segment the four structures: AA, LA-blood, LV-blood, and LV-myo, and compare the segmentation accuracy on the translated images. Fig. \ref{fig:seg_results} shows the qualitative comparison of segmentation results between our SASAN and other methods for CT to MRI image translation, while the quantitative evaluations are shown in Table \ref{table:domain_adaption_results_heart} and Table \ref{table:domain_adaption_results_heart_ct_to_mr} for both adaptation directions. Our code is available on GitHub.\footnote{\url{https://github.com/devavratTomar/sasan}}.


\subsection{Performance Metrics}
We employ the widely used evaluation protocols, the average symmetric surface distance (ASSD) and the Dice similarity coefficient (Dice), to quantitatively measure the performance of domain adaptation models for the segmentation task. In particular, we run a segmentation model on the synthesized medical images to see how well the predicted segmentation map matches ground-truth. The evaluation is conducted based on the subject-level segmentation volume. A higher Dice value and a lower ASSD value indicate more substantial capabilities of adaptation models to preserve the organs’ anatomical structures and generate realistic images. It should also be noted that if one of the segmentation labels is missing, the ASSD values become infinity. In this case, we upper bound the ASSD score to 50.


\subsection{Unsupervised Domain Adaptation}
We evaluate the effectiveness of the proposed method for the cross-domain image synthesis via measuring segmentation accuracy on the fake MRI (synthesized from CT) using a detached U-Net model trained on real MRI. We also observed an improvement in the segmentation accuracy when the detached U-Net is trained with real MRI along with the fake MRI images and the corresponding fake segmentation labels generated by the Attention Module as shown in Table \ref{table:ablation_studies_heart}. We first provide the performance upper bound of supervised training and then obtain the performance lower bound “W/o adaptation” by directly applying the segmentation model learned in the source domain to test target images without using any cross-domain adaptation method. For a fair comparison, we utilize the same network architecture and experimental setup for all baselines experiments. Table \ref{table:domain_adaption_results_heart} reports the segmentation results of unsupervised domain adaptation (MRI $\to$ CT) methods for the cardiac dataset. As shown in Table \ref{table:domain_adaption_results_heart}, when directly applying the trained segmenter on MRI images to test data (CT images), we obtain the Dice score of 0.09$\pm$0.09, indicating that severe domain gap would severely impede the generalization ability of deep models when compared to the performance of upper bound (the Dice score of 0.85$\pm$0.04). Our proposed SASAN outperforms state-of-the-art domain adaptation methods, consistently providing improvements across different modalities for the segmentation performance. For the synthesized MRI from CT images, we improved the Dice score to 0.78$\pm$0.07 over the four cardiac segmentation classes, with the ASSD score being decreased to 4.9$\pm$1.5. In addition, the domain adaptation results for the reverting direction (CT $\rightarrow$ MRI) are shown in a Table \ref{table:domain_adaption_results_heart_ct_to_mr}. Overall, it is difficult to locate the cardiac substructures for the reverting direction and in the synthesized CT from MRI image because of its limited contrast with the surrounding tissue. The qualitative segmentation results in Fig. \ref{fig:seg_results} show that it is challenging to delineate cardiac structure without domain adaptation. Instead, our method enables a flexible adaptation of the segmentation of four cardiac structures. Besides, anatomy-consistency is encouraged with our approach, yielding satisfactory segmentation results of different organs with varying shapes and sizes. Furthermore, We show the visualization of attention maps for the cardiac substructures in Fig. \ref{fig:cardiac_attention}. As indicated by qualitative results, the proposed Attention regularization loss term ensures learning orthogonal attention maps, thus focusing on different anatomical regions and facilitating the translation process across image modalities.

\subsection{Comparison With State-of-The-Art Methods}
We compare our method with three leading unsupervised domain adaptation approaches for the cardiac segmentation: the PnP-Ada-Net model \cite{dou2018pnp}, the CycleGAN model \cite{CycleGAN2017}, and the U-GAT-IT model \cite{kim2019u}, respectively. The U-GAT-IT \cite{kim2019u} is the current state-of-the-art attention-based method for image synthesis. The four cardiac structures (categories), including AA, LA-blood, LV-blood, and LVmyo, are used for segmentation performance evaluation. We summarize the \textit{mean}$\pm$\textit{std} of ASSD and Dice metrics in Table \ref{table:domain_adaption_results_heart} (MRI$\to$CT) and Table \ref{table:domain_adaption_results_heart_ct_to_mr} (CT$\to$MRI). Overall, our SASAN achieves higher Dice score over the non-attention cross-domain adaptation baseline (CycleGAN). Besides, our method surpasses the state-of-the-art attention-based method (U-GAT-IT) by a large margin of 0.21 in Dice and 6.0 in ASSD, respectively. Moreover, as shown in Fig. \ref{fig:seg_results}, translated results using SASAN are visually superior to other methods, e.g., U-GAT-IT, while preserving the source domain's anatomical structures. This highlights the importance of our proposed self-attentive spatial feature normalization, facilitating translation of various tissue regions differently from one domain to another without distortion of the source domain's structures. Finally, except for LA-blood, SASAN performs favorably against one of the state-of-the-art frameworks PnP-Ada-Net. Similar to our approach, PnP-Ada-Net utilized domain adaptation based on output segmentation. However, compared with PnP-Ada-Net, segmentation results generally get improved with our self-attention module that attends to different anatomical structures of the organ, enabling tissue-specific translation.

We also compare the proposed SASAN with SIFA\cite{chen2020unsupervised} using Welch's t-test for the null hypothesis: the average Dice scores are the same for the two methods. As mentioned in Table \ref{table:domain_adaption_results_heart}, for MRI to CT domain adaptation, we get the p-value $<$ 0.3, implying we can reject the above hypothesis with a confidence probability $>$0.7. Similarly, in table \ref{table:domain_adaption_results_heart_ct_to_mr}, we get the p-value$<$0.2 for CT to MRI domain adaptation. We achieve marginal improvement compared to the SIFA method \cite{chen2020unsupervised} in terms of average Dice score.



\begin{table*}[t]
\centering
\caption{Ablation studies for MRI to CT domain adaptation for cardiac substructures segmentation. Abbreviations - w/: with, w/o: without, aug.: data augmentation using fake images and fake labels, Att.: Number of Attention maps, low res.: image resolution of 128$\times$128.}
\resizebox{\linewidth}{!}{%
\begin{tabular}{@{}|c|cc|cc|cc|cc|cc|@{}}
\toprule
\textbf{Method} & \multicolumn{2}{c|}{\textbf{AA}} & \multicolumn{2}{c|}{\textbf{LA-blood}} & \multicolumn{2}{c|}{\textbf{LV-blood}} & \multicolumn{2}{c|}{\textbf{LV-myo}} & \multicolumn{2}{c|}{\textbf{Mean}} \\
                & \textbf{Dice}  & \textbf{ASSD}   & \textbf{Dice}      & \textbf{ASSD}     & \textbf{Dice}      & \textbf{ASSD}     & \textbf{Dice}    & \textbf{ASSD}     & \textbf{Dice}    & \textbf{ASSD}   \\ \midrule
SASAN w/ aug. & 0.82$\pm$0.02 & 4.14$\pm$1.6 & 0.76$\pm$0.25 & 8.3$\pm$2.2 & 0.82$\pm$0.03 & 3.5$\pm$1.1 & 0.72$\pm$0.08 & 3.3$\pm$0.9 & 0.78$\pm$0.10 & 4.9$\pm$1.5\\
SASAN w/o aug.  & 0.70$\pm$0.09  & 20.3$\pm$2.4 & 0.72$\pm$0.11      & 14.6$\pm$5.3 & 0.74$\pm$0.11     & 7.5$\pm$2.4 & 0.68$\pm$0.11 & 7.9$\pm$3.9 & 0.71$\pm$0.07   & 12.6$\pm$2.0    \\\midrule

SASAN w/o $D_{seg}$ & 0.63$\pm$0.11 & 22.3$\pm$2.7 & 0.57$\pm$0.16 & 27.3$\pm$14.9 & 0.71 $\pm$0.04 &  8.2$\pm$2.9 & 0.61$\pm$0.10 & 6.5$\pm$2.0 & 0.63$\pm$0.05 & 16.0$\pm$8.9 \\  

SASAN w/o  $\mathcal{L}_{aux}$ & 0.60$\pm$0.11 & 24.2$\pm$3.3 & 0.55$\pm$0.15 & 26.0$\pm$14.1 & 0.62$\pm$0.04 & 8.9$\pm$3.0 & 0.53$\pm$0.10 & 6.5$\pm$2.4 & 0.58$\pm$0.04 & 16.4$\pm$8.8     
\\ 
  
SASAN w/o  $\mathcal{L}_{reg}$ & 0.61$\pm$0.04 & 24.9$\pm$0.9 & 0.81$\pm$0.05 & 11.1$\pm$5.3 & 0.55$\pm$0.08 & 7.6$\pm$2.4 & 0.53$\pm$0.16  & 10.0$\pm$4.9 & 0.62$\pm$0.11 & 13.5$\pm$6.8  
  \\
SASAN w/ 16 Att. & 0.59$\pm$0.12 & 22.3$\pm$3.6 & 0.58$\pm$0.03 & 23.4$\pm$7.3 & 0.54$\pm$0.13 & 7.9$\pm$1.3 & 0.45$\pm$0.16  & 10.6$\pm$2.8 & 0.54$\pm$0.06 & 16.0$\pm$6.9 
  \\
  
SASAN low res.  & 0.48$\pm$0.14  & 14.10$\pm$1.44 & 0.73$\pm$0.03 & 11.31$\pm$2.97 & 0.55$\pm$0.23 & 4.66$\pm$2.70 & 0.52$\pm$0.22 & 6.14$\pm$1.15 & 0.57$\pm$0.10 & 9.05$\pm$3.82  
  \\
SASAN $\mathcal{L}_{reg}=5$  & 0.68$\pm$0.12  & 20.58$\pm$2.87 & 0.67$\pm$0.15 & 20.44$\pm$10.42 & 0.58$\pm$0.20 & 8.37$\pm$2.78 & 0.64$\pm$0.15 & 9.83$\pm$2.41 & 0.64$\pm$0.04 & 14.80$\pm$5.73  
  \\
SASAN $\mathcal{L}_{aux}=6$.  & 0.67$\pm$0.10  & 21.24$\pm$1.68 & 0.81$\pm$0.05 & 10.68$\pm$3.18 & 0.61$\pm$0.17 & 6.96$\pm$3.62 & 0.66$\pm$0.17 & 6.09$\pm$1.98 & 0.69$\pm$0.07 & 11.24$\pm$6.02  
  \\
\bottomrule
\end{tabular}
\label{table:ablation_studies_heart}
}
\end{table*}

\begin{table}[]
\centering
\caption{Performance comparison for unsupervised cross-domain (MRI-T2 $\rightarrow$ MRI-T1) adaptation of brain segmentation. Abbreviations- Necrosis: Necrotic (fluid-filled) core, NE Tumor: Non-enhancing (solid) core, E Tumor: enhancing core.}
\label{table:domain_adaption_results_brain}
\resizebox{\linewidth}{!}{
\begin{tabular}{@{}|c|c|c|c|c|c|@{}}
\toprule
\textbf{Method} & \textbf{Necrosis} & \textbf{Edema} & \textbf{NE Tumor} & \textbf{E Tumor} & \textbf{Mean} \\ \midrule

CycleGAN & 0.39$\pm$0.10 & 0.55$\pm$0.08 & 0.13$\pm$0.11 &0.40$\pm$0.07 & 0.38$\pm$0.09\\ 
SASAN (Ours) & 0.44$\pm$0.08  & 0.61$\pm$0.07 & 0.18$\pm$0.06 & 0.46$\pm$0.07 & 0.42$\pm$0.07 \\
Upper-bound  & 0.50$\pm$0.07 & 0.65$\pm$0.06 & 0.35$\pm$0.05& 0.51$\pm$0.06 & 0.50$\pm$0.06\\\bottomrule
\end{tabular}}
\end{table}

\subsection{Ablation Studies}

\textbf{Efficacy of Output Segmentation-Level Adaptation.}
As mentioned previously, to further address the domain shift between image modalities, we build an additional convolutional discriminator on top of the auxiliary prediction level obtained by the attention module. We investigate the effectiveness of this output-level adaptation using an ablation experiment for the segmentation task. To do so, we turn-off the adversarial loss term on segmentation by removing the second discriminator and then generating the baseline results for evaluation. We compare this variant (SASAN w/o $D_{seg}$) to our final model, which aligns the joint space of cardiac structures segmentation outputs. As shown in Table \ref{table:ablation_studies_heart}, output space adaptation on cardiac segmentation prediction achieves performance gain of about 0.08 in average Dice score and 3.4 average ASSD score. This indicates that coupling the output-level adaptation (segmentation labels) with the input image-level adaptation improves the performance of segmentation. 

\textbf{Efficacy of Auxiliary Loss.}
We also assess the effects of segmentation loss for the real source domain images (MRI) and the synthetic target domain images (synthetic CT) generated during training. Even though we have segmentation labels for MRI images only, we used the same segmentation labels for the synthetic CT images as they were generated from MRI and thus should have the same semantic layout. As shown in Table \ref{table:ablation_studies_heart}, using these pseudo segmentation labels help the attention module of the CT domain to focus on different cardiac structures of the real CT images.

\textbf{Efficacy of Attention Maps Orthogonality.}
As evident in Table \ref{table:ablation_studies_heart}, we observe that removing redundancy of information from the attention maps is helpful to get better domain adaptation results. We conduct ablation experiments with and without attention regularization loss (SASAN w/o $\mathcal{L}_{reg}$) as described in Section \ref{sec:loss} and observe an overall improvement of 9\% in average Dice score. We also increased the number of attention maps from 8 to 16 and reported the results for each substructure in Table \ref{table:ablation_studies_heart}. Increasing the number of attention maps may add redundancy and decrease performance. We also observe that sometimes without regularization on the attention maps, the attention module tends to focus on one cardiac substructure (LA-blood), leading to slight improvement. However, the overall segmentation accuracy without regularization decreases.

\textbf{Sensitivity Test for the Image Resolutions.}
We also repeated the experiment using different image scaling to evaluate the performance's influence, particularly with an input image resolution of $128\times128$. Overall, as shown in Table \ref{table:ablation_studies_heart}, the results demonstrate, increasing image resolution yields better image translation and higher adaptation performance based on segmentation accuracy.

\subsection{Efficacy of Image Synthesis for Data Augmentation 
}
We show that the use of our SASAN leads to realistic synthetic images that contribute to improving the segmentation performance if synthetic images are used alongside the real images to fine tune the model. In particular, we observed an improvement for the upper bound U-Net model’s accuracy by performing data augmentation with synthetic MRI images and their corresponding ground-truth segmentation labels from the CT domain. Besides, the synthetically generated images can alleviate the common problem of class imbalance despite having a tiny number of real training samples, thus resulting in a more robust system. Table \ref{table:seg_boost} shows the performance gain of SASAN in the mean Dice score to 0.87 using data augmentation with fake MRI when compared to the average Dice score of 0.85 without data augmentation.

Moreover, we also observe an improvement (as shown in Table \ref{table:ablation_studies_heart}) in the mean Dice and ASSD scores when the detached U-Net is trained on the real MRI data with the ground-truth segmentation labels and the synthetic MRI and synthetic labels generated by our method using the generator and attention modules.

\subsection{Domain Adaptation for Brain Tumor Segmentation}
\begin{figure}
    \centering
    \begin{tabular}{@{}c@{}c@{}c@{}c@{}c@{}}
    \includegraphics[width=0.20\linewidth]{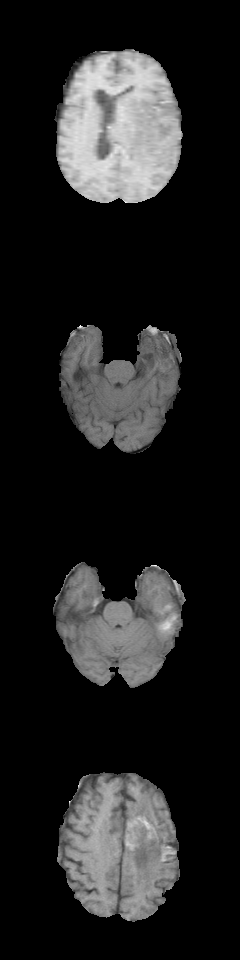} &
    \includegraphics[width=0.20\linewidth]{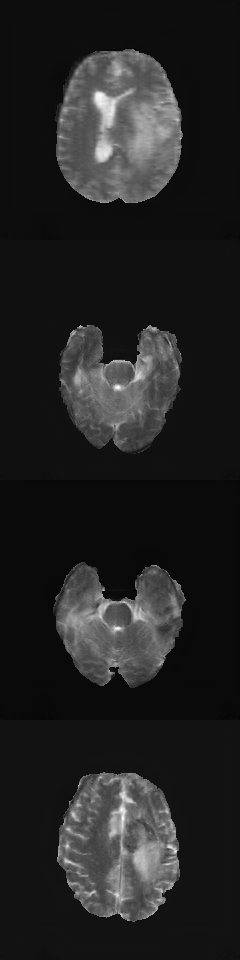} &
    \includegraphics[width=0.20\linewidth]{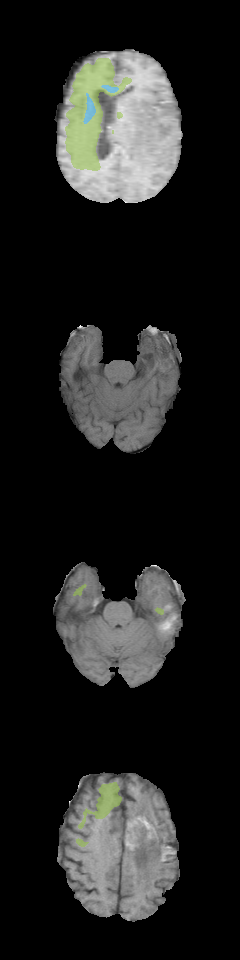} &
    \includegraphics[width=0.20\linewidth]{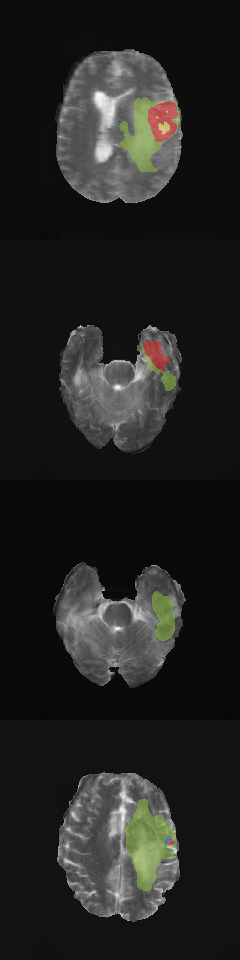} &\includegraphics[width=0.20\linewidth]{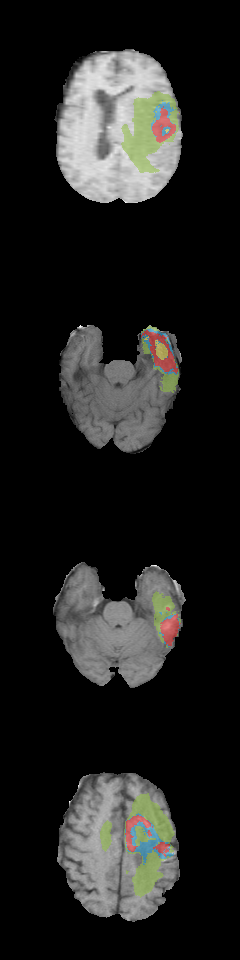}\\
    (a) & (b) & (c) & (d) & (e)
    \end{tabular}
    \caption{Visual comparison of segmentation results produced by domain adaptation from MRI-T2 to MRI-T1: (a) original MRI-T1, (b) fake MRI-T2 (synthesized from MRI-T1) , (c) segmentation results on MR-T1 image using a detached U-Net model trained on real MRI-T2 images, (d) segmentation results with domain adaptation, and (e) ground-truth. We obtain the average Dice score of \textbf{0.42$\pm$0.07} with domain adaptation compared to the Dice score of only 0.02$\pm$0.06 without domain adaptation.}
    \label{fig:t1_t2}
    
\end{figure}

Our unsupervised domain adaptation method can also faithfully translate brain tumor regions from MRI-T1 modality to MRI-T2 modality, as depicted in Fig. \ref{fig:t1_t2}. As the T2 domain has a different contrast, directly feeding T1 brain images to the T2 segmentation model gives poor results with an average Dice score of just $0.02$ for the four tumor classes. On the other hand, our T2 to T1 domain adaptation method increases the average Dice score to $0.42$. Furthermore, we add a comparison with an additional baseline method, CycleGAN, and report the upper-bound for the brain segmentation classes in Table \ref{table:domain_adaption_results_brain}.

\begin{table}
\centering
\caption{Improvement in MRI segmentation accuracy for a model trained on available training MRI data and additional synthetic CT data (CT to MRI translation) with ground-truth.}
\label{table:seg_boost}
\resizebox{\columnwidth}{!}{%
\begin{tabular}{|c|c|c|} 
\toprule
\begin{tabular}[c]{@{}c@{}}\textbf{Cardiac}\\\textbf{ Substructure} \end{tabular} & \begin{tabular}[c]{@{}c@{}}\textbf{Dice w/o}\\\textbf{ Data Augmentation} \end{tabular} & \begin{tabular}[c]{@{}c@{}}\textbf{Dice with}\\\textbf{ Data Augmentation} \end{tabular}  \\ 
\midrule
\textbf{AA}         & 0.84$\pm$0.05 & 0.87$\pm$0.05\\
\textbf{LA-Blood}   & 0.86$\pm$0.08 & 0.86$\pm$0.07\\
\textbf{LV-Blood}   & 0.92$\pm$0.03 & 0.91$\pm$0.03\\
\textbf{LV-myo}     & 0.79$\pm$0.03 & 0.82$\pm$0.03\\ 
\midrule
\textbf{Mean}      & 0.85$\pm$0.04  & 0.87$\pm$0.04\\
\bottomrule
\end{tabular}
}
\end{table}

\section{Experiments: Supervised Setup}

\begin{figure*}[t]
    \centering
    \begin{tabular}{@{}c@{}c@{}c@{}c@{}c@{}c@{}c@{}c@{}c@{}c@{}}
    \includegraphics[width=0.1\linewidth]{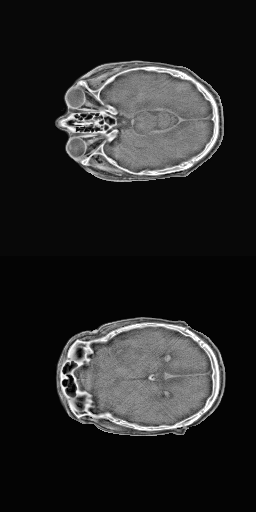} &
    \includegraphics[width=0.1\linewidth]{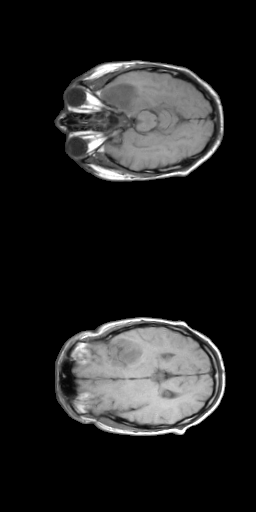} &
    \includegraphics[width=0.1\linewidth]{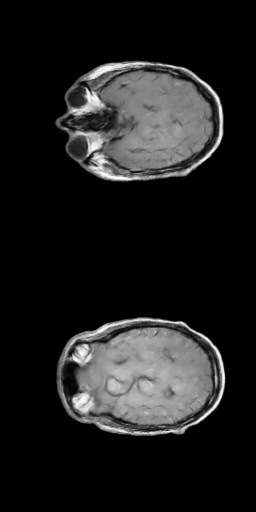} &
    \includegraphics[width=0.1\linewidth]{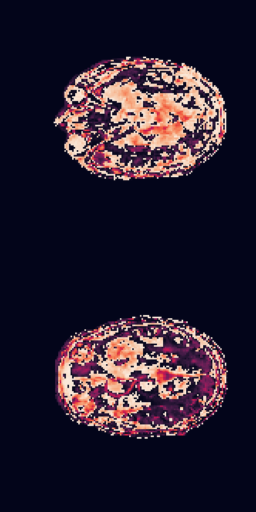} &
    \includegraphics[width=0.1\linewidth]{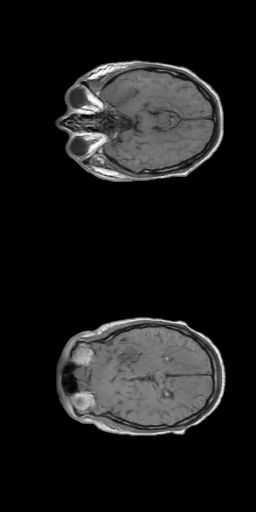} &
    \includegraphics[width=0.1\linewidth]{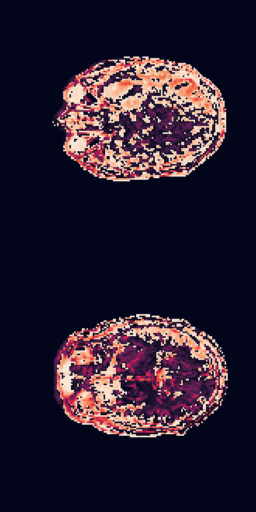} &
    \includegraphics[width=0.1\linewidth]{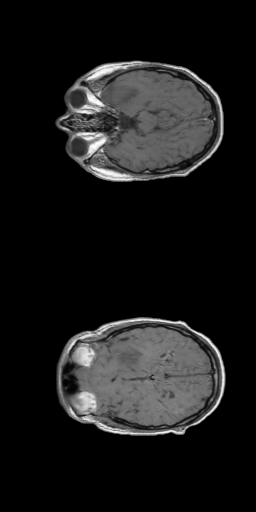} &
    \includegraphics[width=0.1\linewidth]{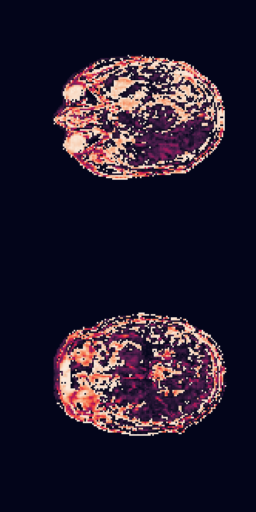} &
    \includegraphics[width=0.1\linewidth]{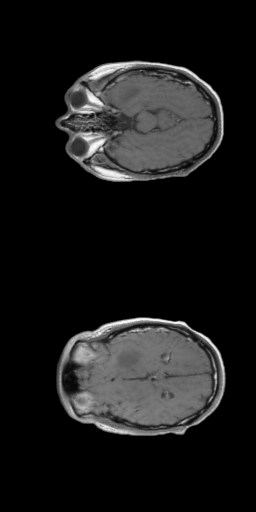} &
    \includegraphics[width=0.1\linewidth]{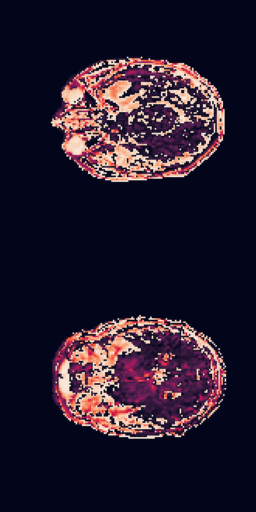}  \\
    (a) & (b) & \multicolumn{2}{c}{(c)} & \multicolumn{2}{c}{(d)} & \multicolumn{2}{c}{(e)} & \multicolumn{2}{c}{(f)} 
    \end{tabular}
    \caption{Qualitative results of different supervised image modality conversion methods followed by respective error maps (brighter color indicates a higher error). From left to right: (a) input CT, (b) reference MRI, the translation results of (c) Pix2Pix, (d) MR-GAN, (e) DC2Anet, (f) SASAN. Zoom in for the best view.}
    \label{fig:rire_results}
\end{figure*}

\subsection{Dataset}
We assess the efficacy of our method for supervised modality conversion for the CT-MR brain dataset \cite{rirewebsite}.

\subsubsection{RIRE Dataset \cite{rirewebsite}} 
This dataset consists of paired but unregistered CT and MR images of 20 patients. We first register the same patient's CT and MRI volumes by maximizing their joint histogram mutual information. Furthermore, we remove the outlier noise by clipping the pixel values between 0.5 and 99.5 percentiles and also mask out the outer circle noise from CT images. We split the data randomly at the patient-level into on-overlapping training (18 patients), and test (2 patients) sets.

\begin{figure}[h]
    \centering
    \begin{tabular}{@{}c@{}c@{}c@{}c@{}}
    \includegraphics[width=0.25\linewidth]{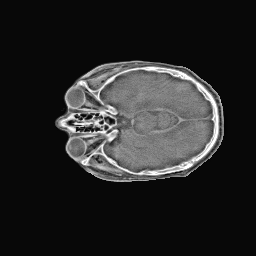} &
    \includegraphics[width=0.25\linewidth]{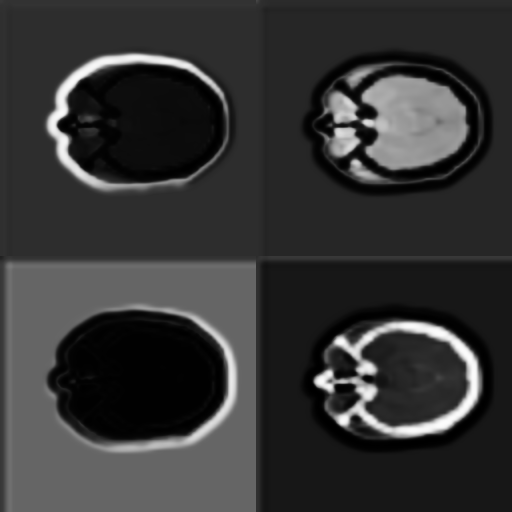} &
    \includegraphics[width=0.25\linewidth]{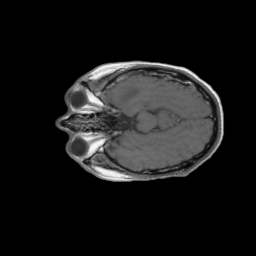} &
    \includegraphics[width=0.25\linewidth]{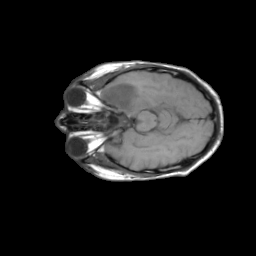}\\
    \includegraphics[width=0.25\linewidth]{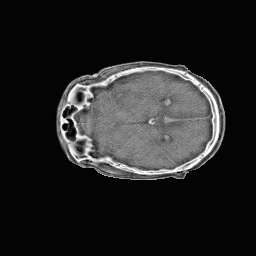} &
    \includegraphics[width=0.25\linewidth]{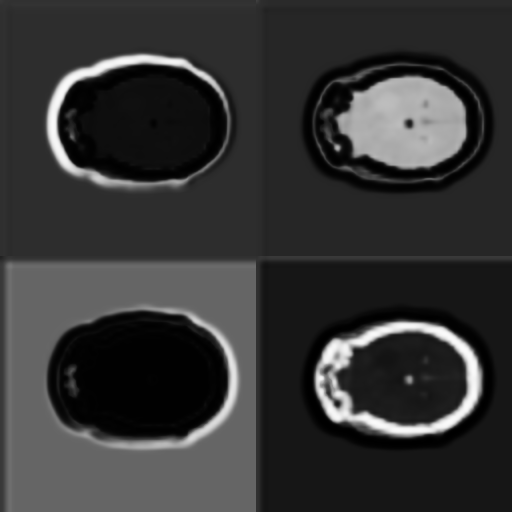} &
    \includegraphics[width=0.25\linewidth]{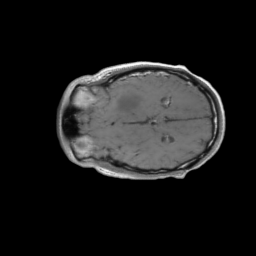} &
    \includegraphics[width=0.25\linewidth]{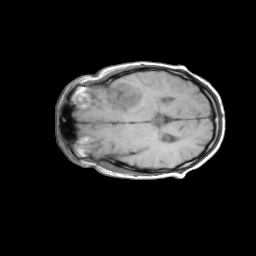}\\
    (a) & (b) & (c) & (d)
    \end{tabular}
    \caption{Illustration of the learned attention maps: (a) original CT images, (b) learned attention maps, (c) generated MRI images, (d) ground-truths of MRI.}
    \label{fig:rire_atten}
\end{figure}

This section presents experimental results of supervised training for image translation on MRI/CT brain images from the RIRE dataset. Our proposed SASAN has been evaluated alongside the baseline supervised methods Pix2Pix, MR-GAN, and DC2Anet. Since the segmentation labels of the RIRE dataset are not available at training, learned attention results are even more critical since they allow us to spare segmentation annotations and capture different complex relationships between tissues/parts of the brain. Fig. \ref{fig:rire_atten} shows sample four attention maps produced by SASAN. It is noticeable that the attention module manages to distinguish between different regions such as inner region, outer frontal contour, outer back contour, etc. This part can also be seen as automation of methods similar to tissue-segmentation based classical approaches. SASAN does not require heavy and expensive annotations of segmentation. However, it manages to refine different regions based on learned attention maps. Regarding the relative performance, the attention module allowed us to improve upon other prominent approaches, as shown in Table \ref{table:rire_results}.

For this, we conducted an evaluation study using the distortion metrics, e.g., the Structural Similarity Index (SSIM), the Peak Signal to Noise Ratio (PSNR), Mean Absolute Error (MAE), Root-Mean-Square Error (RMSE), Pearson Correlation Coefficient (PCC) as well as the learned perceptual image patch similarity (LPIPS) \cite{zhang2018unreasonable} for the supervised setting. The purpose of this experiment is to evaluate how synthesized MRI images recover realistic textures faithful to bone/tissue regions. 

As shown in Table \ref{table:rire_results}, Our SASAN achieves competitive performance with respect to supervised image translation-based methods. However, as shown in \cite{wang2018recovering,rad2019srobb}, distortion metrics such as the PSNR used as quantitative metrics do not match up directly with perceptual quality. For example, the synthesized MRI images using both our SASAN and the DC2Anet are not ranked first in terms of the PSNR or MAE metrics; however, they generate more visually appealing results in Fig. \ref{fig:rire_results} than the other methods, and they can preserve the underlying geometry and structure of anatomical regions. This shows leveraging the geometric cues; our model produces realistic textures and sharper edges for synthesized MRI images, which are not easily visible on CT images. In addition, we added an error map between the generated image and the reference image for each baseline method

\begin{table}[ht!]
\centering
\caption{Performance comparison with different supervised image translation methods (CT $\rightarrow$ MRI) on the RIRE dataset. }
\label{table:rire_results}
\resizebox{\columnwidth}{!}{%
\begin{tabular}{@{}lllllll@{}}
\toprule
             & \textbf{LPIPS} & \textbf{MAE} & \textbf{PSNR} & \textbf{SSIM} & \textbf{RMSE} & \textbf{PCC} \\ \midrule
Pix2Pix      &  0.06$\pm$0.01 & 0.02$\pm$0.01 & 22.42$\pm$0.97 & 0.87$\pm$0.01 & 0.08$\pm$0.01 & 0.96$\pm$0.01  \\
MR-GAN       &  0.05$\pm$0.01 & 0.03$\pm$0.01 & 21.46$\pm$0.74 & 0.90$\pm$0.02 & 0.08$\pm$0.01 & 0.97$\pm$0.01\\
DC2Anet      &  0.05$\pm$0.01 & 0.04$\pm$0.01 & 20.99$\pm$0.98 & 0.90$\pm$0.01 & 0.09$\pm$0.01 & 0.97$\pm$0.01 \\
SASAN (Ours) &  0.05$\pm$0.01    &      0.04$\pm$0.01        &     21.37$\pm$1.29          &    0.90$\pm$0.02           &     0.09$\pm$0.01          &       0.97$\pm$0.01       \\ \bottomrule
\end{tabular}
}
\end{table}

\section{Conclusion and Future Work}
This paper proposed a generic domain adaptation framework, SASAN, to synthesize missing modality for biomedical images and improve the generalization ability of CNNs across different image modalities. The proposed approach is based on a self-attention mechanism that attends various anatomical structures of the organ in the image to improve cross-domain image synthesis. Besides, our proposed auxiliary loss can be used in a plug-and-play way to leverage semantic information and guide the attention network, thus facilitating the transformation. Promising results in brain image synthesis and cardiac segmentation show the effectiveness of our approach. Further ablation studies have validated that SASAN improves existing synthesis methods in quantitative and qualitative measures, establishing new state-of-the-art.

The current limitation of unsupervised domain adaptation methods is they need a fully labeled source domain, which hinders their applicability in the practical clinical workflow. A possible future extension of this work would be utilizing self-supervised learning to leverage unlabeled data, yielding more practical domain adaptation, where only a small fraction of data in source domain is annotated.

\bibliographystyle{IEEEbib}
\bibliography{references}

\section{Appendix}
\subsection{Implementation Details}
The SASAN network architecture builds upon several building blocks, such as ConvBlock (reflection pad followed by Conv2d and ReLU) and SPADEResnetBlock.

\begin{figure}[ht]
  \centering
  \includegraphics[width=\linewidth]{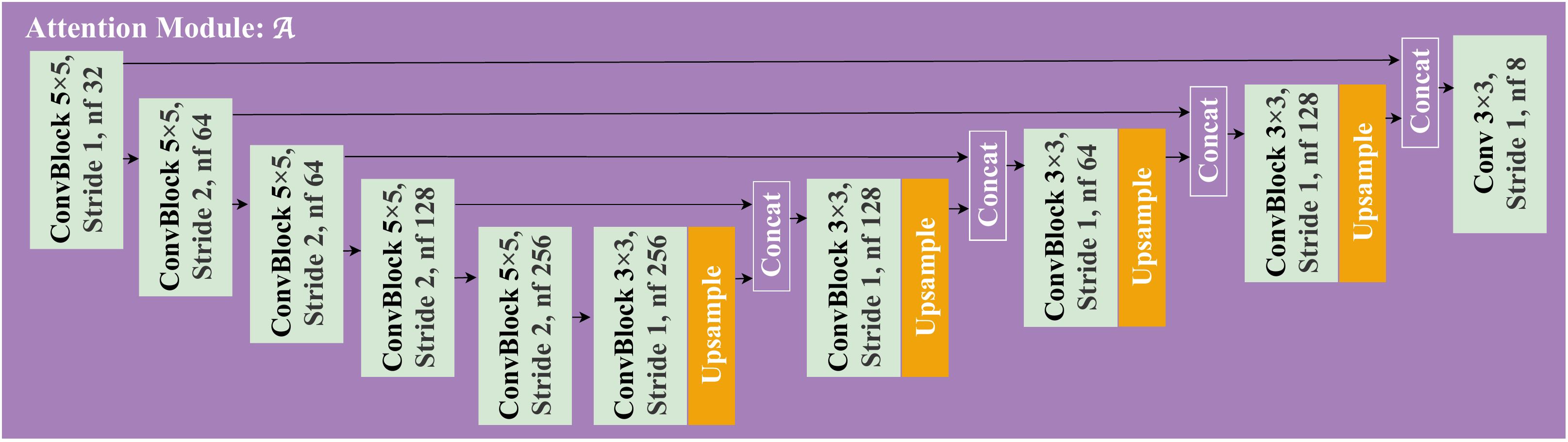} 
  \caption{The architecture of the Attention Module.}
\label{fig:attention_module}
\end{figure}

The generator makes use of encoder-decoder architecture (encoder followed by a decoder). Encoder consists of four Convolution blocks with output channels equal to $32, 64, 128, 128$, kernel size - $7\times7, 3\times3, 3\times3, 3\times3$, stride - $1, 2, 2, 1$ and reflection padding - $3, 1, 1, 1$. Hence, given $3$ input features of $256\times256$ encoder outputs $128$ output features in $64\times64$ dimensions. On the other hand, the Decoder consists of three SPADEResnetBlocks with input features equal to $128, 128, 64$, number of attentions - $8$ and output features - $128, 64, 32$, respectively. Two Convolution blocks then follow SPADEResnetBlocks in Decoder with the number of output channels equal to $16, 3$ respectively, kernel size - $3\times3$, stride - $1$, and padding - $1$. Since SPADEResnetBlocks apply on three space resolutions, we upsample (using bilinear mode and scale factor of $2$) the output of the first two of them. Fig. \ref{fig:attention_module} shows the architecture of the Attention Module.

Our model consists of two PatchGAN based discriminators \cite{isola2017image}, one for segmentation and another for images. Discriminator of images starts with a Convolutional layer having $32$ output channels, kernel size - $4\times4$, stride - $2$ and padding - $1$. The convolutional layer is then followed by LeakyReLU ($0.2$) and two additional Conv2D with output channels equal to $64, 128$, kernel size - $4\times4$, stride - $2$ and padding - $1$. The output of each of the last two convolutional layers goes through InstanceNorm2d and LeakyReLU ($0.2$) modules. Finally, the image discriminator concludes with a convolution layer with $1$ output channel, kernel size - $4\times4$, stride - $1$ and padding - $1$. Discriminator for segmentation is almost identical, the only difference being the number of output channels in the middle two convolutional layers equal to $32$. Thus both patch-based discriminators have a field of view of $8\times8$.

We implement the U-Net based segmentation model consisting of four down-sampling layers (Maxpool along with DoubleConv -$2\times$ of Conv, BatchNorm, and ReLU modules) followed by four up-sampling ones (upsampling with factor 2 along with DoubleConv). Additionally, the segmentation module contains an initial DoubleConv prior to down-sampling and final Conv2d after up-sampling.

\subsection{Hyper-parameters}
We found experimentally that the following hyper-parameters works the best for the losses defined in section \ref{methods}:
$\lambda_c = 10.0,  \lambda_{id} = 2.5, \lambda_{reg} = 1.0, \lambda_{aux} = 0.1
$. We use Adam\cite{kingma2014adam} optimizer with a learning rate of $10^{-4}$ and $\beta_1=0.5$, $\beta_2=0.999$. The whole model is trained end-to-end using the batch size of $2$.  
\end{document}


\title{Supplementary Material: Self-Attentive Spatial Adaptive Normalization for Cross-Modality Domain Adaptation}
\author{Devavrat Tomar, Manana Lortkipanidze, \\ 
Behzad Bozorgtabar, and Jean-Philippe Thiran, \IEEEmembership{Senior Member, IEEE} 

\thanks{D. Tomar. is with the Signal Processing Laboratory (LT55), \'Ecole Polytechnique F\'ed\'erale de Lausanne (EPFL), Lausanne, Switzerland (e-mail: devavrat.tomar@epfl.ch).}

\thanks{M. Lortkipanidze is with the Signal Processing Laboratory (LT55), \'Ecole Polytechnique F\'ed\'erale de Lausanne (EPFL), Lausanne, Switzerland (e-mail: manana.lortkipanidze@epfl.ch)}

\thanks{B. Bozorgtabar is with the Signal Processing Laboratory (LT55), \'Ecole Polytechnique F\'ed\'erale de Lausanne (EPFL), Lausanne, Switzerland, and with the Radiology Department, Centre Hospitalier Universitaire Vaudois, Lausanne, Switzerland, and also with the Centre d\'Imagerie Biom\'edicale (CIBM) (e-mail: behzad.bozorgtabar@epfl.ch).}
\thanks{J. Thiran is with the Signal Processing Laboratory (LT55), \'Ecole Polytechnique F\'ed\'erale de Lausanne (EPFL), Lausanne, Switzerland, and with the Radiology Department, Centre Hospitalier Universitaire Vaudois, Lausanne, Switzerland, and with the Centre d\'Imagerie Biom\'edicale (CIBM), and also with the University of Lausanne, Lausanne, Switzerland (e-mail: jean-philippe.thiran@epfl.ch).}}

\maketitle



\section{Other Results}
    \label{sup:baselines}
    
\bibliographystyle{IEEEtran}
\bibliography{references.bib}